\documentclass{article}

\usepackage[preprint]{neurips_2026}

\usepackage[utf8]{inputenc}
\usepackage[T1]{fontenc}
\usepackage{hyperref}
\usepackage{amsmath,amsfonts,amssymb,amsthm}
\usepackage{graphicx}
\usepackage{algorithm2e}
\usepackage{stmaryrd}
\usepackage{dsfont}
\usepackage{xfrac}
\usepackage{mathrsfs}
\usepackage{xcolor}

\SetKwComment{Comment}{/* }{ */}

\newcommand{\mma}{{\textsc{MMA}}}
\newcommand{\tomato}{{\textsc{ToMATo}}}
\newcommand{\tomatomp}{{\textsc{ToMAToMP}}}
\newcommand{\dgm}{{\rm Dgm}}

\newcommand{\R}{\mathbb{R}}
\newcommand{\Xn}{\hat{X}_n}
\newcommand{\clustering}{\mathcal{C}}
\newcommand{\im}{\mathrm{im}}
\newcommand{\matching}{\mathrm{m}}
\newcommand{\linefam}{\mathcal{L}}
\newcommand{\interval}{\mathcal{I}}
\newcommand{\eps}{\varepsilon}
\newcommand{\outliers}{\mathcal{O}}
\newcommand{\slicedfunc}[2]{F_{#1}(#2)}
\newcommand{\pbar}{\mathcal B}
\newcommand{\prom}{{\rm prom}}

\title{\tomatomp: Robust and Multi-Parameter Topological Clustering}

\author{
\textbf{Ludo Andrianirina}
\quad
\textbf{Mathieu Carri\`ere}$^{1}$\thanks{Corresponding author: \texttt{mathieu.carriere@inria.fr}}\\[1mm]
$^{1}$DataShape, Centre Inria d'Université Côte d'Azur, Sophia-Antipolis, France\\
}

\newtheorem{theorem}{Theorem}
\newtheorem{corollary}{Corollary}
\newtheorem{definition}{Definition}

\begin{document}
\maketitle

%--------------------------------------------------
\begin{abstract}
Topological clustering, and its main algorithm \tomato, is a clustering method from Topological Data Analysis (TDA) which has been applied successfully in several applications during the last few years. This is due to its high versatility, as clusters are detected from the persistent components in the sublevel sets of any user-defined function (gene expression, pixel values, etc), and efficiency, as topological clustering enjoys robustness guarantees. However, \tomato\ is also limited in several ways. First, a graph on the data points needs to be provided as a hyper-parameter of the method (whose fine-tuning is left to the user). Second, \tomato\ is known to be very sensitive to outlier values in the function range. Finally, and most importantly, \tomato\ can only handle one function at a time, whereas it is critical to use several functions in various applications. \textbf{In this article, we introduce \tomatomp: the first topological clustering method able to handle several functions at the same time with theoretical guarantees.} More specifically, we leverage a recent tool from multi-parameter persistent homology, called MMA decomposition, to design our clustering algorithm, and prove that it enjoys robustness properties. As corollaries, we show that it can be used to make \tomato\ independent of graph tuning, and robust to outliers. Finally, we provide a set of numerical experiments showcasing the efficiency and quality of the clusterings produced by \tomatomp, by showing strong improvement over non-topological and topological baselines for various datasets.
\end{abstract}

%--------------------------------------------------
\section{Introduction}

In the field of unsupervised learning, the \emph{topological clustering} method, originally introduced in~\cite{chazalPersistenceBasedClusteringRiemannian2013}, has recently become increasingly relevant in various application scenarios. Intuitively, this method belongs to the family of hierarchical clustering methods, where the hierarchy is built on top of the outputs of \emph{mode-seeking} algorithms, which aim at finding the \emph{modes}, or \emph{basins of attraction}, of a density estimator $f$ (such as, e.g., Mean Shift~\cite{chengMeanShiftMode1995} or~\cite{koontzGraphTheoreticApproachNonparametric1976}).

However, contrary to traditional mode-seeking algorithms, which suffer from several caveats including, e.g., sensitivity to initialization or noise, the modes obtained with topological clustering are captured using the toolset of \emph{persistent homology} (PH),
which not only allows to build a hierarchy between modes, but also ensures theoretical robustness of the computed clusters, which usual hierarchical clustering algorithms typically lack (as illustrated, for instance, by the empirical instability of dendrograms~\cite{carlssonCharacterizationStabilityConvergence2010}).

Moreover, while topological clustering was introduced with density estimators to mimic the framework of mode-seeking algorithms, the modes of many other functions have been introduced and used in the last few years, each related to the tasks and datasets at hand: Heat Kernel Signature for segmenting 3D shapes~\cite{skrabaPersistenceBasedSegmentationDeformable2010}, smoothed gene expression for quantifying spatial variations~\cite{boyleTopologicalDataAnalysis2026}, gene expression gradient for segmenting tissues~\cite{steinAccurateTilingSpatial2025}, pixel values for capturing morphological features~\cite{chungMorphologicalMultiparameterFiltration2024}, etc. This illustrates the high versatility of topological clustering, which has proved efficient in very different applications.

The main topological clustering algorithm is called \tomato~\cite{chazalPersistenceBasedClusteringRiemannian2013} (though several variants exist in the literature, such as \textsc{AuToMATo}~\cite{huberAuToMAToOutOfTheBoxPersistenceBased2024} or \textsc{STopover}~\cite{baeSTopoverCapturesSpatial2025}), and builds a hierarchy by processing the data points in decreasing order of the function values, which can be thought of as a scan of the dataset; each time a local maxima is encountered, a corresponding mode has been found. Moreover, these modes are progressively filled in by looking at the connected components that \emph{persist} as the scanning of the data points goes on: after its detection with its local maxima, the mode will progressively grow until the scan reaches a saddle point, which connect the mode to another one. This interval (from the mode detection to its merge) is called the \emph{lifetime}, or \emph{prominence}, of the mode, and quantifies its importance. The merging also allows to build a hierarchy (by recording which modes gets merged to which), which, in turn, allows to only keep the modes that pass a given prominence threshold (similarly to cutting a dendrogram at a given height). See Figure~\ref{fig:tomato}

\paragraph{Limitations of \tomato.} However, \tomato\ suffers from a few weaknesses. First, deciding whether a data point is a regular point, a saddle, a local minimum or a local maximum requires to build a graph $G$ on top of the data points (point types are then assessed based on the neighbor values). Such graph is a hyper-parameter whose tuning is left to the user, despite being critical for the clustering quality. Second, while the PH framework allows to guarantee that \tomato\ clusters are robust to small function perturbations, \tomato\ is sensitive to \emph{outlier} values: only a few points with aberrant function values are sufficient to destroy \tomato\ clusters. Finally, there are many applications where the modes of \emph{several} functions are relevant (for instance, pairs or triplets of gene expressions), but \tomato\ can only handle one function at a time.

\paragraph{Contributions.} In this article, we propose \tomatomp: the first (to our knowledge) multi-parameter topological clustering algorithm with theoretical guarantees. In particular:

\begin{enumerate}
	\item[$(i)$] We leverage the recent framework of \emph{multi-parameter PH}, and, in particular, \emph{MMA decompositions}~\cite{loiseauxMultiparameterModuleApproximation2025}, to design a new multi-parameter topological clustering algorithm \tomatomp\ (Algorithm~\ref{alg:tomatomp}). Our code can be found here: \texttt{https://github.com/MathieuCarriere/tomatomp}. % \textcolor{red}{TODO}
	\item[$(ii)$] We prove that \tomatomp\ enjoys a robustness guarantee, similar to the one of \tomato\ (Theorem~\ref{thm:tomatomp}).
	\item[$(iii)$] As direct corollaries, we prove that \tomatomp\ allows to $(a)$ use \tomato\ without any graph parameter tuning, and $(b)$ make \tomato\ robust against outliers (Corollaries~\ref{cor:metric} and~\ref{cor:outliers}), hence fixing two critical limitations of \tomato.
	\item[$(iv)$] We demonstrate the relevance of our method on several applications, ranging from 3D shape segmentation to gene rankings in spatial transcriptomics datasets, with favorable results against standard topological and non-topological baselines.
\end{enumerate}

\paragraph{Related work.} While topological clustering is still a recent field, 
%using (multi-parameter) PH and/or 
extending standard topological clustering (with or without PH) has been studied in various contexts. For instance, Huber et al.~\cite{huberAuToMAToOutOfTheBoxPersistenceBased2024} have proposed bootstrap methods for automatically tuning the prominence threshold in order to capture only those statistically significant modes. Kim and Mémoli~\cite{kimExtractingPersistentClusters2023} have used the framework of multi-parameter PH to handle and quantify the evolution of dynamic hierarchical clusterings when datasets change over time.
More closely related to our work, Rolle and Scoccola~\cite{, rolleStableConsistentDensityBased2024, scoccolaPersistablePersistentStable2023} have proposed to study the stability of \emph{slices} (i.e., reductions of several functions into one using, e.g., linear combinations) of multi-parameter PH in order to design robust one-parameter topological clusterings. While this approach provides nice stability guarantees, it also requires the user to pick a slice, which can be seen as a new hyper-parameter. On the other hand, we emphasize that our proposed \tomatomp\ method %(while also based on slices of multi-parameter PH), 
has essentially the same hyperparameters than \tomato\ 
%(except when one function is used to avoid graph parameter tuning---in that case, it has one parameter less) 
and thus does not require to pick a specific slice; instead, it leverages the whole multi-parameter PH structures.

\paragraph{Plan.} In Section~\ref{sec:background}, we recall the basics of \tomato\ and (multi-parameter) PH, with a focus on MMA decompositions. Then, in Section~\ref{sec:algorithm}, we introduce our new algorithm \tomatomp, and present its 
theoretical properties in Section~\ref{sec:theory}. Finally, numerical experiments are showcased in Section~\ref{sec:experiments}, and concluding remarks are provided in Section~\ref{sec:conclusion}.

%--------------------------------------------------
\section{Background}\label{sec:background}

In this section, we recall the basics of \tomato\ and (multi-parameter) PH. See  Appendix~\ref{app:background} and Algorithm~\ref{alg:tomato} for more details, and~\cite{edelsbrunnerComputationalTopologyIntroduction2010, oudotPersistenceTheoryQuiver2015, deyComputationalTopologyData2022} for thorough presentations. We first provide an (informal) description of \tomato, and then connect it to (multi-parameter) PH.

\begin{figure}[h!]
	\centering
	\includegraphics[width=9.5cm]{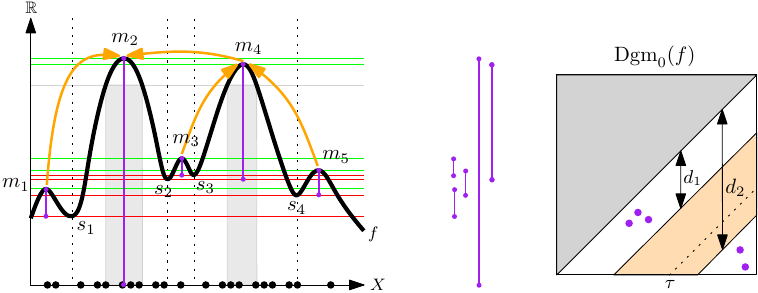}
	%\includegraphics[width=5cm]{volcano.png}
	%\includegraphics[width=5cm]{synthetic-clusters.png}
	%\caption{Persistence in dimension $0$ scales almost linearly, and produces a diagram (left) on which one can clearly see that two points have much larger prominences than the rest (including one with infinite persistence displayed in green): they correspond to the two spirals, which, after running persistence again using a threshold $\tau$ corresponding to the prominence separating those two points from the rest, provides a clean clustering (right).}
	\caption{Example of \tomato\ clustering induced by a density estimator $f$ defined on a space $\Xn\subset\R$. The different local maxima $m_i$ are connected together (with orange arrows) at the saddles $s_i$, and prominences are displayed as purple bars. The corresponding PD is displayed on the right: every mode is associated to a PD point, and the PD is $(d_1,d_2)$-separated. Choosing a threshold $d_1\leq\tau\leq d_2$ induces a final clustering with two clusters.}
	\label{fig:tomato}
\end{figure}

\paragraph{Persistence-Based Clustering: \tomato.} %\label{sec:tomato}
The main idea of \tomato\ is to compute clusters as the modes, or basins of attraction, of a continuous function $f$, and to use \emph{persistence theory} to build a hierarchy between clusters.
Roughly speaking, the algorithm works by taking as input a graph $G$ whose vertex set $V(G)$ is the dataset $\Xn$. Then, points in $\Xn$ are sorted in decreasing function order, and processed in different ways depending on the point type (which is assessed with the point neighbors in $G$): local maxima induce the creation of a new mode that starts persisting, while saddles and local minima induce the merging of two modes, and the end of one of the two modes persistence; as the modes with lowest maxima are always merged to the mode with highest maximum, this induces a hierarchy between modes. Each mode is represented with a point $p\in\R^2$, using the function values of the corresponding local maxima and saddle/local minima, called the \emph{birth} and \emph{death} times respectively; their difference is called the \emph{prominence} $\prom(p)$ of the mode. %is called the  (resp. \emph{death time}) of the mode, 
Such points are stored in a so-called \emph{persistence diagram} $\dgm_0(f)$. %and their 
%the absolute difference is called the \emph{prominence}, and the hierarchy is encoded in a so-called \emph{persistence diagram} $\dgm_0(f)$. 
See Figure~\ref{fig:tomato}. 

\paragraph{Notations.} Given a dataset $\Xn$, a function $\Xn\to\R$, a graph $G$, and a prominence threshold $\tau > 0$, we let $\clustering_\tomato:\Xn\to\llbracket 1,N\rrbracket$ (where $N$ is the number of clusters) be the corresponding clustering $\clustering_\tomato=\tomato(\Xn, f, G, \tau)$. %Moreover, we assume, unless otherwise stated, that the PD points are indexed in a consistent way, i.e., $\dgm_0(f)=\{(b_i,d_i)\}_{1\leq i\leq M}$, where $b_i$ (resp. $d_i$) is the birth (resp. death) time of cluster $i$.

\paragraph{Single-Parameter PH.} The persistence diagram (PD) introduced in the \tomato\ algorithm is the main descriptor of \emph{persistent homology} (PH), which studies the persistence of topological features (encoded in homology groups) in general spaces. %filtrations, i.e., sequences of growing subspaces. See Appendix~\ref{app:background} for more details.
Several robustness properties have been established for PDs, and directly apply to \tomato. They are stated using \emph{diagram distances} (including the so-called \emph{bottleneck distance} $d_b$), which are similar to optimal transport distances, as they are obtained by computing the cost of matching PD points together (and to the diagonal if PDs have different numbers of points). %between PDs.
The exact definition of such distances is not necessary for the main exposition, and we provide it in Appendix~\ref{app:background}. %The most common diagram distance is the ., which enjoys stability.

%Importantly, the \tomato\ algorithm enjoys some \emph{stability} and \emph{correctness} guarantees. The stability property ensures that the PD is stable w.r.t. function perturbations.

%\begin{theorem}[\cite{cohen-steinerStabilityPersistenceDiagrams2007}]
%Let $\Xn$ be a dataset, and $f,g:\Xn\to\R$ be two functions defined on it. 
%Let $\dgm_0(f), \dgm_0(g)$ be their corresponding PDs computed by \tomato\ using the same graph $G$.
%Then,
%\begin{equation}
%	d_b(\dgm_0(f), \dgm_0(g)) \leq \|f-g\|_\infty.
%\end{equation}
%where $d_b$ stands for the \emph{bottleneck distance} between the PDs.
%\end{theorem}

The robustness property of \tomato\ requires that 
%ensures that the clustering computed by \tomato\ is correct, provided that 
the PD points are sufficiently separated: %, as in the following definition.

\begin{definition}[{\cite[Definition 4.1]{chazalPersistenceBasedClusteringRiemannian2013}}]\label{def:d12-sep}
	Let $0\leq d_1 < d_2$. The PD $\dgm_0(f)$ is called \emph{$(d_1,d_2)$-separated} if its points have prominences either less than $d_1$ or more than $d_2$.
	See Figure\ref{fig:tomato}.
\end{definition}

%\begin{theorem}[{\cite[Theorem 9.2]{chazalPersistenceBasedClusteringRiemannian2013}}]
%	Let $\Xn\subset X$ be a dataset of $n$ points sampled i.i.d. on an $m$-dimensional manifold $X$ with positive convexity radius $\rho(X)$, and let $f$ be a $c$-Lipschitz probability density function w.r.t. the $m$-dimensional Hausdorff measure with $(d_1,d_2)$-separated PD	 $\dgm_0(f)$. Finally, let $G_\delta$ be a $\delta$-neighborhood graph computed with the geodesic distance on $X$,
%	with $\delta\in ]0, \min\{ \rho(X), \frac{d_2-d_1}{5c}\}[$.
%	Let $\tau \in ]d_1+2c\delta, d_2-3c\delta[$.
%	Then $\clustering_{\tomato}:=\tomato(\Xn,f,G_\delta,\tau)$ recovers the correct number of clusters (i.e., equal to the number of modes of $f$) with probability $1-\exp(-\Omega(n))$.
%\end{theorem}

%See Figure~\ref{fig:tau_choice}. Moreover, stability can be improved for separated PDs, as it is possible to define correspondences between \tomato\ clusters of sufficiently close functions.

\begin{theorem}[{\cite[Theorem 10.1]{chazalPersistenceBasedClusteringRiemannian2013}}, {\cite[Theorem 84]{rolleStableConsistentDensityBased2024}}]\label{thm:stab-tomato-clusters}
	Let $0 \leq d_1 < d_2$.
	Let $\Xn$ be a dataset, and $f,g:\Xn\to\R$ be two functions defined on it
	s.t. $\dgm_0(f)$ is $(d_1,d_2)$-separated and
	$\eps := \|f-g\|_\infty \leq \frac{d_2-d_1}{16}$. Then,
	%\begin{equation}
		$d_b(\dgm_0(f), \dgm_0(g)) \leq \eps$. %\|f-g\|_\infty.
	%\end{equation}
	
	Moreover, let $\clustering_f:=\tomato(\Xn, f, G, \tau)$ and $\clustering_g:=\tomato(\Xn, g, G, \tau)$, with $d_1 < \tau < d_2$.
	Then, $\clustering_f$ and $\clustering_g$ are mutually \emph{$3\eps$-related}:
	there exists a bijection $b : \im(\clustering_f)\to\im(\clustering_g)$ s.t., for every cluster $C_i=\clustering_f^{-1}(i)$, $i\in \im(\clustering_f)$, with associated PD point $p_i=(b_i,d_i)\in\dgm_0(f)$, one has $C_i\cap f^{-1}([d_i+3\eps,+\infty[)\subseteq \clustering_g^{-1}(b(i))$ (and vice-versa).
\end{theorem}

Note that while \tomato\ was described for a (finite) dataset $\Xn$, it can be applied straightforwardly (and PDs as well) to a continuous space $X$ and function $f:X\to\R$ (in this case, there is no need to provide a graph $G$). We let $\tomato(X,f,\tau)$ denote the corresponding clustering.

%--------------------------------------------------
%\subsection{(Multi-Parameter) Persistence and MMA Descriptor}

\paragraph{Multi-parameter PH.} It is known %in the Topological Data Analysis literature 
that generalizing single-parameter PH (i.e., PH with a single function) to the setting where \emph{several} functions $f_1,\dots,f_p$, $p>1$, are provided is very difficult, as 
%there is no decomposition theorem (into indicator modules)
%ensuring the existence 
there is no analogue of PDs for multi-parameter PH. 
Hence, several approaches have turned to \emph{slicing}, i.e., to returning to single-parameter PH using 
%by turning the different functions into a single one. A robust approach is to do it with 
\emph{diagonal lines}.

\begin{definition}
	Let $X$ be a topological space. Let $f_1,\dots,f_p\to\R$ be continuous functions. Let $\ell$ be a diagonal line in $\R^p$, i.e., a line with direction $[1,\dots,1]$, and let
	$\varphi :\R\to\ell$ be a linear parametrization of $\ell$.
	The function $\slicedfunc{\ell}{f_1,\dots,f_p}:X\to\R$ is defined as the function %whose sublevel sets satisfy: 
	satisfying:
	$\slicedfunc{\ell}{f_1,\dots,f_p}^{-1}([t,+\infty[)=\bigcap_{i=1}^p f_i^{-1}([[\varphi(t)]_i, +\infty[)$\footnote{Recall that $[\cdot]_i$ denotes the $i$-th coordinate in $\R^p$.}. Moreover, given %for any point
	 $p_i=(b_i,d_i)\in\dgm_0(\slicedfunc{\ell}{f_1,\dots,f_p})$, we let the \emph{bar} $\pbar(p_i)$ denote the segment in $\R^p$ with endpoints $\varphi(b_i), \varphi(d_i)$. 
\end{definition}

Note that the previous definition also holds in the discrete setting (dataset $\Xn$, graph $G$).

Using bars from functions obtained from diagonal lines allows to build the so-called \emph{MMA decompositions}. These decompositions aim at mimicking PDs for multi-parameter PH: they are comprised of shapes in $\R^p$, each shape being interpreted as the lifetime of a connected component.

\begin{definition}
	Let $X$ be a topological space.
	Let $f_1,\dots,f_p:X\to\R$ be continuous functions. 
	Let %$\fdgm_0(f_1,\dots,f_p; \linefam)$ be the corresponding fibered PD for a given 
	$\linefam$ be an ordered family of diagonal lines $\linefam=\{\ell_1,\dots,\ell_{|\linefam|}\}$. 
	Moreover, let $\matching_{\ell_i}$ be a \emph{matching} function, i.e., an 
	injective function between $\dgm_0(\slicedfunc{\ell_i}{f_1,\dots,f_p})$ and $\dgm_0(\slicedfunc{\ell_{i+1}}{f_1,\dots,f_p})$.
	%the PDs associated to $\ell_i$ and $\ell_{i+1}$, 
	%that is, an injective function $\matching_{\ell_i}:
	%mapping the points of  $
	%\dgm(\ell_i,f_1,\dots,f_p)
	%$ to a subset of the points of $
	%\to\dgm(\ell_{i+1}, f_1,\dots,f_p)$.
	The \emph{\mma\ decomposition} is defined as:
	\begin{equation}
		\mma_0(f_1,\dots,f_p; \linefam, \{\matching_{\ell}\}_{\ell\in\linefam}):=\{\interval(S)\}_{S\in \mathcal D(\{\matching_{\ell}\}_{\ell\in\linefam})},
	\end{equation}
	where $\mathcal D(\{\matching_{\ell}\}_{\ell\in\linefam})$ denotes the set of sets of  consecutive bars matched together under the matching functions, that is, each $S$ is of the form $S:=\{b_{i_1},\dots,b_{i_{|S|}}\}$ where $b_{i_j}=\pbar(p)$ for some $p\in \dgm_0(\slicedfunc{\ell_{i_j}}{f_1,\dots,f_p})$ and $b_{i_{j+1}}=\pbar(\matching_{\ell_{i_j}}(p))$ for all $1\leq j\leq |S|-1$, and where $\interval(S)$ denotes an interval\footnote{Recall that an interval in $\R^p$ is a connected union of hyperrectangles, i.e., Cartesian products of intervals in $\R$.} with smallest volume that one can form using the endpoints of the bars in $S$.
\end{definition}

\begin{figure}[h!]
	\centering
	\includegraphics[width=0.8\textwidth]{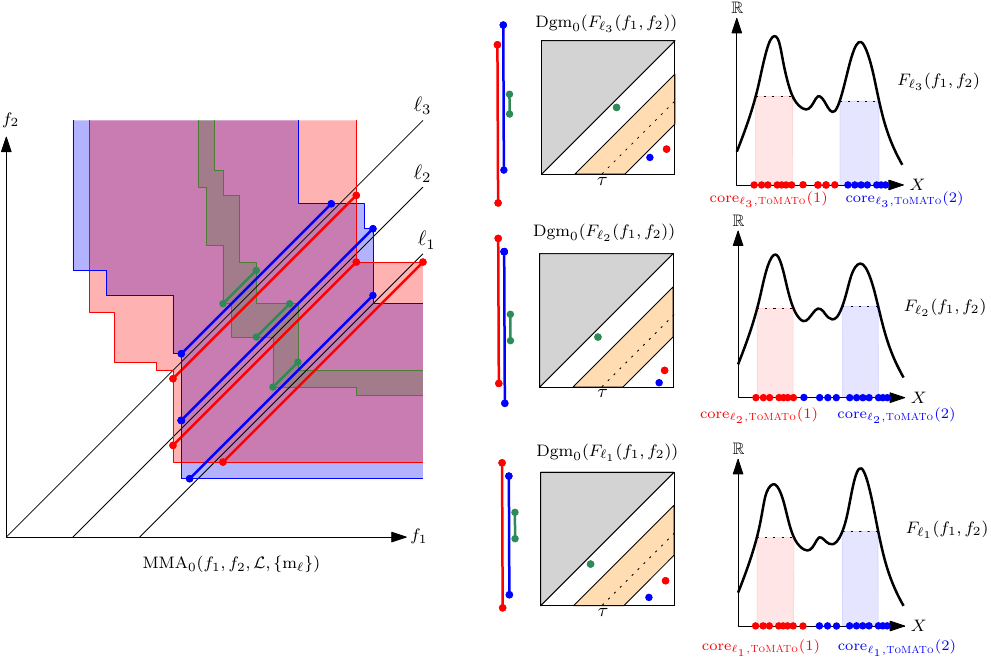}
	\caption{{\textbf{(Left)}} MMA decomposition with induced PDs of three diagonal lines, displayed both with points in $\R^2$ and bars displayed along the lines. {\textbf{(Right)}} Each PD (obtained by running \tomato\ with $\slicedfunc{\ell}{f_1,f_2}$) leads to a clustering of the dataset. The cores of the clusters (computed as the subparts of the clusters that are sufficiently above the saddles) remain stable across lines.}
	\label{fig:mma}
\end{figure}

Note that MMA decompositions can also be computed in the discrete setting (dataset $\Xn$, graph $G$).
See Figure~\ref{fig:mma}.
Note also that, given some diagonal line $\ell$ that does not necessarily belongs to the family $\linefam$ used to compute an MMA decomposition $\mma=\{\mathcal I_i\}_{1\leq i \leq N}$, there is a corresponding \emph{induced PD} $\mma|_\ell:=\{\mathcal I_i\cap \ell\}_{1\leq i \leq N}$. By construction, on lines belonging to $\linefam$, induced PDs coincide with the PDs of the $\slicedfunc{\ell}{f_1,\dots,f_p}$s: $\mma|_\ell=\{\pbar(p)\}_{p\in\dgm(\slicedfunc{\ell}{f_1,\dots,f_p})}, \forall\ell\in\linefam$.

\section{\tomatomp: Algorithm and Guarantees}\label{sec:tomatomp}

In this section, we introduce \textbf{\tomatomp}, the first multi-parameter topological clustering method based on \mma\ decompositions. We first provide our algorithm in Section~\ref{sec:algorithm}, and then state our theoretical guarantees in Section~\ref{sec:theory}.

\subsection{The  \tomatomp\ algorithm}\label{sec:algorithm}
The main idea is quite simple: start with a family of diagonal lines $\linefam$, compute the \tomato\ clusterings on every line, relate the \tomato\ clusters of different lines together using the intervals appearing in the MMA decomposition (so that all \tomato\ clusters are now indexed by the MMA intervals), %---see Algorithm~\ref{alg:lineclustering}), 
and, for every data point $x$, assign it to the most frequent \tomato\ cluster that it belongs to across the lines. See Algorithm~\ref{alg:tomatomp}. Its complexity is driven by the one of computing \mma\ decompositions, which is cubic in the number of graph nodes and edges.

\begin{algorithm}[h!]
	\caption{The \tomatomp\ algorithm}\label{alg:tomatomp}
	\KwData{$\Xn, f_1,\dots,f_p, G, \tau, \linefam, \{\matching_\ell\}_{\ell\in\linefam}$}
	\KwResult{$\clustering_{\tomatomp}$}
	%\For{$\ell\in\linefam$}{
		%	$\dgm_0(\ell,f_1,\dots,f_p),\clustering_{\tomato}\leftarrow\tomato(\Xn,(\ell,f_1,\dots,f_p),G,\tau)$\;
		%}
	$\{\interval_i\}_{1\leq i \leq N}\leftarrow \mma_0(f_1,\dots,f_p;\linefam,\{\matching_\ell\}_{\ell\in\linefam})$\;
	\For{$\ell\in\linefam$}{
		$\clustering_{\ell,\tomato}\leftarrow \tomato(\Xn,\slicedfunc{\ell}{f_1,\dots,f_p},G,\tau)$\;
		$\pi_\ell\leftarrow$ permutation %of 
		%$\llbracket 1,N \rrbracket\to\llbracket 1,N \rrbracket$ 
		s.t. $b_{i}\in \mma|_{\ell} = \pbar(p_{\pi_\ell(i)}), p_{\pi_\ell(i)}\in\dgm_0(\slicedfunc{\ell}{f_1,\dots,f_p})$\;
		\For{$x\in\Xn$}{
			$\clustering_{\ell,\mma}(x)\leftarrow \pi_\ell(\clustering_{\ell,\tomato}(x))$\;
		}
		%$\clustering_{\ell,\mma}\leftarrow\textsc{CreateClustering}(\{\interval_i\}_{1\leq i \leq N},\ell,G,\tau)$\;
	}
	\For{$x\in\Xn$}{
		$\clustering_{\tomatomp}(x)\leftarrow {\rm argmax}_{1\leq i \leq N}\ |\{ \ell\in\linefam \,:\, \clustering_{\ell,\mma}(x) = i\}|$\;
	}
	\Return{$\clustering_{\tomatomp}$}
\end{algorithm}

\subsection{Theoretical Robustness of \tomatomp}\label{sec:theory}

We now state and prove the main guarantee associated to \tomatomp. In short, our main result states that, if the input has well-separated induced PDs (and thus clear distinction between noisy and relevant clusters for every line), then the clusters produced by \tomatomp\ will be in bijection with the corresponding PD points of the induced PDs with large prominences. See Figure~\ref{fig:mma}.

\begin{theorem}[Robustness of \tomatomp]\label{thm:tomatomp}
Let $X$ be a topological space.
Let $f_1,\dots,f_p:X\to\R$ be continuous functions, 
%Let $f_1,\dots,f_p:\Xn\to\R$ be continuous functions defined on the data space, 
and let $G$ be some graph defined on the data points.
Let $\linefam=\{\ell_1,\dots,\ell_{|\linefam|}\}$ be an ordered family of diagonal lines at distance $\eta > 0$, and let $\{\matching_\ell\}_{\ell\in\linefam}$ be a family of compatible matching functions\footnote{That is, the pairs of bars matched together by each $\matching_\ell$ have no strictly comparable endpoints: the start points $s_1,s_2$ corresponding to bars $b_1,b_2$ s.t. $b_2=\matching_\ell(b_1)$ do not satisfy $s_1<_p s_2$ or $s_2<_p s_1$ (and similarly for endpoints), where $x\leq_p x'\in\R^p$ iif $[x]_i\leq [x']_i$ for all $1\leq i\leq p$.}  achieving the $q$-th diagram distances between consecutive PDs.
\begin{enumerate}
	\item[(A1)] Assume that $\exists 0\leq d_1 < d_2$ s.t. $\forall \ell\in\linefam$, $\dgm_0(\slicedfunc{\ell}{f_1,\dots,f_p})$ is
	$(d_1,d_2)$-separated. 
	%F(\ell,f_1,\dots,f_p)$ share the same modes (as defined with $G$), and that the corresponding PD points all have prominences at least $d_2$, 
	%Assume that $\dgm_0(\ell,f_1,\dots,f_p)$ is $(d_1,d_2)$-separated for all $\ell\in\linefam$, 
	\item[(A2)] Let $d^*:=\frac 12 \min\{\|p-p'\|_\infty:\exists\ell\in\linefam \text{  s.t. }p,p'\in \dgm_0(\slicedfunc{\ell}{f_1,\dots,f_p}), \prom(p) \geq d_2, \prom(p')\geq d_2\}$. 
	%\text{ for some }\ell\in\linefam\text{ and } p,p'\text{ have prominence at least }d_2\}$. 
	Assume $\eta \leq \min\{\frac{d_2-d_1}{16},d^*\}$.
\end{enumerate}
Then:
\begin{enumerate}
	\item[$(i)$] the matching functions $\{\matching_\ell\}_{\ell\in\linefam}$  are unique and restrict to bijections between points of consecutive PDs with prominences at least $d_2$,
	\item[$(ii)$]  for any $d_1 < \tau < d_2$, 
	%if $\eta \leq \min\{\|p-p'\|_\infty\,:\,p,p'\in \dgm_0(\ell,f_1,\dots,f_p)\text{ for some }\ell\in\linefam\}$, then 
	the number of clusters $N$ computed by 
	$\clustering_{\tomatomp}:=\tomatomp(\Xn, f_1,\dots,f_p, G, \tau, \linefam, \{\matching_\ell\}_{\ell\in\linefam})$ 
	is equal to the number $N$ of 
	points of $\dgm_0(\slicedfunc{\ell}{f_1,\dots,f_p})$ with prominences at least $d_2$\footnote{This number $N$ is the same for all $\ell\in\linefam$ thanks to $(i)$.}, and
	\item[$(iii)$] Given a cluster index $1\leq i\leq N$, let $S_i:=\{b_i^\ell %=(b_i^\ell,d_i^\ell)
	\}_{\ell\in\linefam}$ denote the set of consecutive bars matched together by the matching functions $\matching_\ell$.
	Then, 
	letting $\clustering_{\ell,\tomato}:=\tomato(\Xn,\slicedfunc{\ell}{f_1,\dots,f_p},G,\tau)$, the family of clusterings $\{\clustering_{\ell,\tomato}\}_{\ell\in\linefam}$ is $3\eta$-related to $\clustering_{\tomatomp}$, i.e., 
	there exists a cluster $C_i=\clustering_{\tomatomp}^{-1}(i)$ s.t.
	$\bigcap_{\ell\in\linefam}
	{\rm core}_{\ell,\tomato}(i) 
	%\left(\clustering_{\ell,\tomato}^{-1}(i)\cap %F(\ell,f_1,\dots,f_p)^{-1}([d^\ell_i + 3\eta,+\infty[)\right) 
	\subseteq C_i$, where the \emph{core} is defined as ${\rm core}_{\ell,\tomato}(i) := \clustering_{\ell,\tomato}^{-1}(\pi_\ell(i))\cap \slicedfunc{\ell}{f_1,\dots,f_p}^{-1}([\varphi_\ell^{-1}(b^\ell_i) + 3\eta, +\infty[)$, where each $\varphi_\ell$ is a linear parametrization of $\ell$.
\end{enumerate} 

%clusters of each \tomato\ clusterings along the lines of $\linefam$, and, for every 
%recovers the correct number of clusters (i.e., equal to the number of modes of the $F(\ell,f_1,\dots,f_p)$).
\end{theorem}

See Figure~\ref{fig:mma} for an illustration of the notations and statements, and  Appendix~\ref{app:proofs} for the proof and a few extensions. %of Theorem~\ref{thm:tomatomp} can be found in .
One may wonder whether $\bigcap_{\ell\in\linefam} {\rm core}_{\ell,\tomato}(i)=\emptyset$, making $(iii)$ trivial. %While this can happen in general, 
Note that choosing $\eta$ sufficiently small while keeping $|\linefam|$ upper bounded guarantees (using Theorem~\ref{thm:line-stab} in Appendix~\ref{app:background}) that the intersection is not empty, as PD points can move at most $\eta$ from a line to another. The local maxima of modes must thus stay in the cores if there are not too many lines. %See also Appendix~\ref{app:proofs} for a few extensions of Theorem~\ref{thm:tomatomp}.
%See also Appendix~\ref{app:proofs} for more details about the proof (and the proofs of Corollaries~\ref{cor:metric} and~\ref{cor:outliers} as well).

%--------------------------------------------------
\subsection{Applications to \tomato}

In this section, we use our main result Theorem~\ref{thm:tomatomp} to break the two aforementioned limitations of \tomato, i.e., the need for the user to provide and/or fine-tune an input graph $G$, and the high sensitivity to outliers in function values.
See Appendix~\ref{app:proofs} for more details.
%Theorem~\ref{thm:tomatomp} has two direct applications.

\paragraph{Neighborhood graph tuning.}
The most frequent graphs used by \tomato\ are $\delta$-neighborhood graphs. As proved in a recent result~\cite[Theorem 3.6 $(iii)$]{andreEstimatingPersistentHomology2025}, it turns out that letting $\delta$ increase across a specific range of values, %can be 
%used as a separate function, which, when 
combined with an input function $f$, %and fed to \tomatomp, 
allows to recover the \tomato\ clusters associated to $f$ without the need to pick a specific $\delta$ using \tomatomp.

%The first application is that \tomatomp\ can be used to avoid choosing the neighborhood graph parameter $\delta$ for \tomato. Indeed, it has been shown recently that varying $\delta$ and a function $f$ simultaneously induce a $2$-parameter persistence module that satisfies the assumptions of Theorem~\ref{thm:tomatomp}.

%\begin{theorem}[{\cite[Theorem 3.6 $(iii)$]{andreEstimatingPersistentHomology2025}}]\label{thm:mp-sfa}
%	Let $\Xn$ be a geodesic $\eps$-sample of a compact geodesic space $X$ with convexity radius $\rho(X)$ s.t. $\eps \leq \rho(X)/4$, let $\delta_{\rm max}\in[4\eps, \rho(X)[$, and let $G:=G_{\delta_{\rm max}}$ be the $\delta_{\rm max}$-neighborhood graph built on $\Xn$. Now, let $f$ be a $c$-Lipschitz function s.t. $\dgm_0(f)$ (computed on $X$) is $(d_1,d_2)$-separated. Finally, let $\tilde G$ be the first barycentric subdivision of $G$, let $\tilde f:V(\tilde G)\to\R$ be the extension of $f$ to $\tilde G$, and let $g:V(\tilde G)\to\R$ be the function assigning $0$ to vertices in $V(G)$ and the lengths of the corresponding edges to vertices in $V(\tilde G)\setminus V(G)$.	
%	Then, the PDs $\dgm_0(\ell,g,\tilde f)$ (computed on $G$) are $(d_1+c\delta_{\rm max}, d_2-c\delta_{\rm max})$-separated for all $\ell\in\linefam$.
%\end{theorem}

\begin{corollary}\label{cor:metric}
	Let $\Xn$ be a geodesic $\eps$-sample of a compact geodesic space $X$ with convexity radius $\rho(X)$ s.t. $\eps \leq \rho(X)/4$, let $\delta_{\rm max}\in[4\eps, \rho(X)[$, and let $G:=G_{\delta_{\rm max}}$ be the $\delta_{\rm max}$-neighborhood graph built on $\Xn$. Now, let $f:X\to\R$ be a $c$-Lipschitz function s.t. $\dgm_0(f)$ is $(d_1,d_2)$-separated. Finally, let $\tilde G$ be the first barycentric subdivision of $G$\footnote{That is, the graph obtained by splitting every edge of $G$ into two edges.}, let $\tilde f:V(\tilde G)\to\R$ be the restriction of $f$ to $V(\tilde G)$, and let $g:V(\tilde G)\to\R$ be the function assigning $0$ to vertices in $V(G)$ and the lengths of the corresponding edges of $G$ to vertices in $V(\tilde G)\setminus V(G)$.
	%Under the same assumptions of Theorem~\ref{thm:mp-sfa}, and 
	Let $\linefam$ be an ordered family of diagonal lines at distance $\eta\leq \frac{d_2-d_1-4c\delta_{\rm max}}{16}$, $d_1+2c\delta_{\rm max}\leq \tau\leq d_2-2c\delta_{\rm max}$, and $\{\matching_\ell\}_{\ell\in\linefam}$ as in Theorem~\ref{thm:tomatomp}.
	Then, $\clustering_{\tomato}:=\tomato(X,f,\tau)$ is $3(c\delta_{\rm max}+\eta)$-related to $\clustering_{\tomatomp}(\Xn,g,\tilde f,\tilde G,\tau,\linefam, \{\matching_{\ell}\}_{\ell\in\linefam})$. 
	%, one has that $\clustering_{\tomatomp}:=\tomatomp(V(\tilde G), g,\tilde f, \tilde G, \tau, \linefam, \{\matching_\ell\}_{\ell\in\linefam})$
	%recovers the cores of the clusters of $f$ (in the sense of Theorem~\ref{thm:tomatomp}).
\end{corollary}

\paragraph{Outlier robustness.}
Another application of Theorem~\ref{thm:tomatomp} is the setting where input functions are plagued with outlier values. More specifically,
if one is given some  function $\tilde f$ that differs from $f$ on a few points, but such that the difference between $f$ and $\tilde f$ is very large on these points, the modes of $f$ and $\tilde f$ might potentially be very different. 
%destroy the \tomato\ clusters as these are stable w.r.t. small perturbations but not to extreme, outlier values. 
In this case, using an outlier detection score as a separate function and combine it with $\tilde f$ allows \tomatomp\ to recover the \tomato\ clusters associated to $f$, provided that the points with outlier function values are \emph{topologically robust}. %used to create a $2$-parameter persistence module for \tomatomp.

\begin{definition}
	Let $\Xn$ be a dataset, and $G_\delta$ be a $\delta$-neighborhood graph built on top of $\Xn$.
	A point $x\in\Xn$ is called \emph{topologically robust} if it has at least one neighbor $n(x)$ in $G$ that is connected to all of its other neighbors.
\end{definition}

\begin{corollary}\label{cor:outliers}
	Let $\Xn$ be a geodesic $\eps$-sample of a compact geodesic space $X$ with convexity radius $\rho(X)$ s.t. $\eps \leq \rho(X)/4$, let $\delta\in[4\eps, \rho(X)[$, and let $G:=G_{\delta}$ be the $\delta$-neighborhood graph built on $\Xn$. Now, let $f:X\to\R$ be a $c$-Lipschitz function s.t. $\dgm_0(f)$ is $(d_1,d_2)$-separated. Moreover, assume one is given another function $\tilde f$ s.t. there exists a set $\outliers\subset V(G)$ of non-overlapping\footnote{I.e., the sets of neighbors are disjoint.}, topologically robust points s.t. $f(x)=\tilde f(x)$ for $x\in V(G)\setminus \outliers$. Assume one is given an outlier score function $g:\Xn\to\R$ s.t.
	$g(x)\geq\max_{y\in\Xn\setminus\outliers}\ g(y)$, for all $x\in\outliers$. 
    %Then, 
	%the PDs $\dgm_0(\ell,g,\tilde f)$ (computed on $G_\delta$) are $(d_1+2c\delta, d_2-2c\delta)$-separated for all $\ell\in\linefam$.
	%As such, picking 
	Let $\linefam$ be an ordered family of diagonal lines at distance $\eta\leq \frac{d_2-d_1-4c\delta}{16}$, $d_1+2c\delta\leq \tau\leq d_2-2c\delta$, and $\{\matching_\ell\}_{\ell\in\linefam}$ as in Theorem~\ref{thm:tomatomp}.
	Then, $\clustering_{\tomato}:=\tomato(X,f,\tau)$ is $3(c\delta+\eta)$-related to $\clustering_{\tomatomp}(\Xn,g,\tilde f|_{\Xn},G,\tau,\linefam, \{\matching_{\ell}\}_{\ell\in\linefam})$. 
	%one has that $\clustering_{\tomatomp}:=\tomatomp(\Xn, g, \tilde f, G, \tau, \linefam, \{\matching_\ell\}_{\ell\in\linefam})$
	%recovers the cores of the clusters of $f$ (in the sense of Theorem~\ref{thm:tomatomp}).
\end{corollary}

In practice, when outliers are not known a priori, one can always ensure that the assumptions of Corollary~\ref{cor:outliers} are satisfied by manually adding edges to $G_\delta$ for all points whose function values are above a given threshold, which can typically be set as a quantile of the function value distribution. Note however that this also turn $G_\delta$ into another graph that is not necessarily a neighborhood graph.

%--------------------------------------------------
\section{Experiments}\label{sec:experiments}

%In this section, we apply our algorithm \tomatomp\ to several datasets. 
%in order to show its usefulness.
%More specifically, we illustrate Corollary~\ref{cor:metric}, Corollary~\ref{cor:outliers} and Theorem~\ref{thm:tomatomp} in Sections~\ref{sec:exp-metric},~\ref{sec:exp-outliers} and~\ref{sec:exp-tomatomp}, respectively.

\begin{figure}[h!]
	\centering
	\includegraphics[width=4cm]{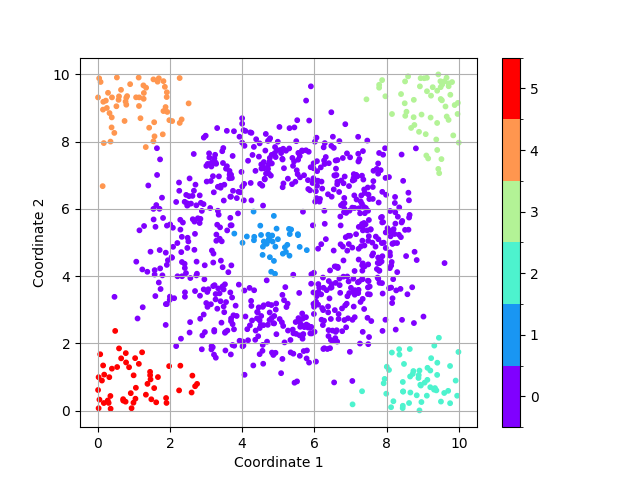}
	\includegraphics[height=3cm]{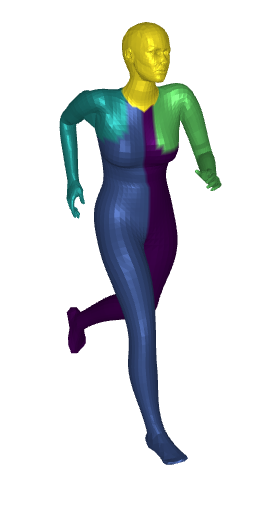}
	\includegraphics[width=4cm]{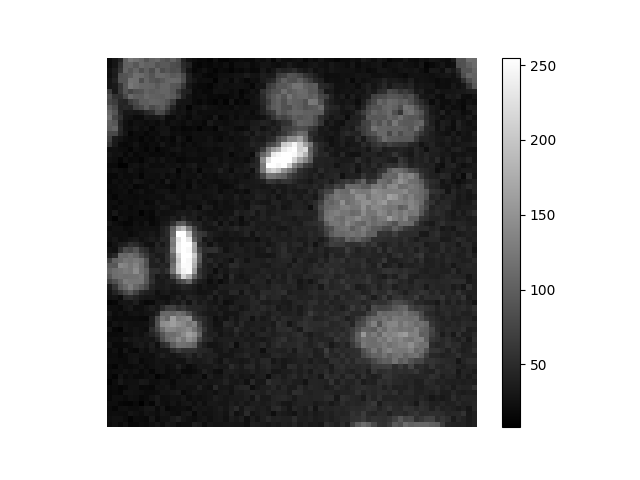}
	\includegraphics[width=4cm]{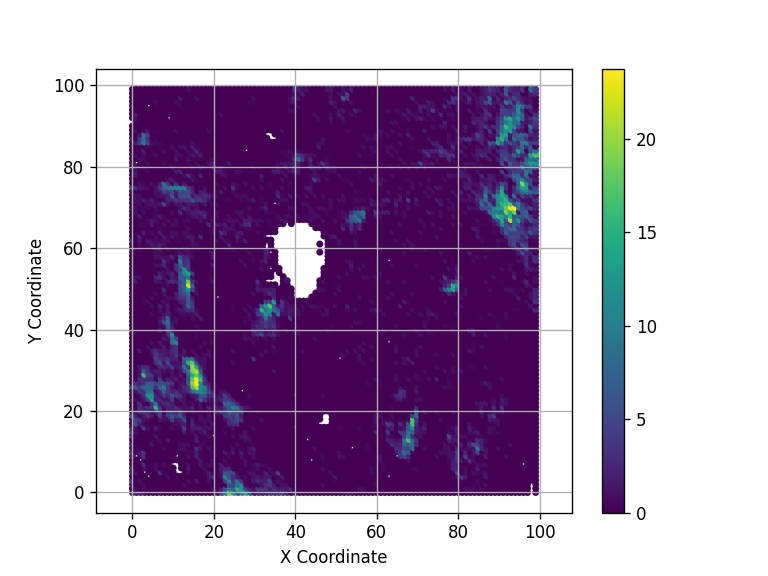}
	\caption{Datasets used in experiments. From left to right: synthetic data, 3D shape, morphological image of cells, and one spatial transcriptomics datasets (spat).}
	\label{fig:datasets}
\end{figure}

%\paragraph*{Datasets.}
In this section, we provide numerical experiments with \tomatomp. We study four use cases of topological clustering: %with a specific function is known to be useful: 
$(i)$ a synthetic dataset in $\R^2$ with the Distance-To-Measure (DTM)~\cite{chazalGeometricInferenceProbability2011} ($n=1,000$ points), $(ii)$ a 3D shape from the Princeton database\footnote{\url{https://segeval.cs.princeton.edu/}} with the Heat Kernel Signature ($t=1,000$)~\cite{skrabaPersistenceBasedSegmentationDeformable2010} ($n=568$ points), $(iii)$ a morphological image representing cells from the Bio-image Analysis Notebooks suite\footnote{\url{https://haesleinhuepf.github.io/BioImageAnalysisNotebooks/intro.html}} with image pixel values ($n=70\times 70$ pixels), and $(iv)$ two spatial transcriptomics datasets, kpmp and spat ($n=645$ points and $n=977$ points respectively) from~\cite{boyleTopologicalDataAnalysis2026} and~\cite{baeSTopoverCapturesSpatial2025}, containing cells in two pieces of tissue with the expression of several genes. See Figure~\ref{fig:datasets}.

\paragraph{Tasks and scores.} On datasets $(i)$-$(iii)$, the goal is to recover the right number of clusters, and Adjusted Rand Index (ARI) and Average Mutual Information (AMI) are used as quality scores. On dataset $(iv)$, the goal is to rank (pairs of) genes based on how many clusters they induce in data, a quantity called \emph{coefficient of spatial structure} (CoSS)~\cite{boyleTopologicalDataAnalysis2026}. For single genes, this coefficient is measured as the squared sum of the prominences of \tomato\ clusters induced by gene expression, while for pairs of genes, we use the average Jaccard similarities between the \tomato\ clusters of both genes, following a known method for detecting co-localization~\cite{baeSTopoverCapturesSpatial2025}. 
Concerning single genes (1-g), we rank the top 50 genes with largest variances, while for pairs of genes (2-g), we rank all pairs from the top 10 genes with largest variances (i.e., 45 pairs).
Two scores are used to measure a ranking quality against ground-truth ranking\footnote{In both datasets in $(iv)$, the ground-truth ranking is obtained by ordering (pairs of) genes with the CoSS obtained after running \tomato\ without outlier values and using a fine-tuned $\delta$ parameter for the neighborhood graph. Hence, as the ground-truth is intrinsically topological, we expect non-topological baselines to fail.}: Pearson correlation and TopHits@10, which measures the fraction of top 10 hits that are also top 10 in the ground-truth ranking. 

\paragraph{Computation details.} All results are presented for a family of $|\linefam|=100$ lines, evenly spaces within the function ranges (note that these ranges are rescaled in applications $(iv)$). Matching functions are computed with the vineyards algorithm~\cite{cohen-steinerVinesVineyardsUpdating2006} (as they achieve the $q$-th diagram distances for sufficiently close lines), and the prominence threshold is either set so as to keep only the relevant number of clusters (when this number is known a priori, as in datasets $(i)$-$(iii)$), or to $\tau=0$ (in datasets $(iv)$). Finally, for application $(iv)$, we design a multi-parameter criterion for ranking (pairs of) genes by computing a quantile ($q=10\%$) of the distribution of the (1-parameter) CoSS's obtained from the induced PDs produced by \tomatomp. As for the non-topological baselines, we focus on hierarchical clustering, and compute the squared sum of distances in the dendrogram (using both cell coordinates and gene expression to compute the distances) as ranking criterion.
%the squared sum (over the clusters produced by \tomatomp) of the quantile of the longest diagonal bar 
In Appendix~\ref{app:add-lines}, we also provide plots showing the quality scores and running times dependence on the number of lines used in \tomatomp. Our code is based on \texttt{multipers}~\cite{loiseauxMultipersMultiparameterPersistence2024}, \texttt{gudhi}~ (\url{https://gudhi.inria.fr/python/latest/}) and \texttt{scikit-learn}~\cite{scikit-learn}, and was run on a cluster with Intel Xeon E5-2620 v3 CPUs with 6 cores and 128 GB RAM.

\paragraph*{Neighborhood graph tuning.}%\label{sec:exp-metric}
In this first set of experiments, we measure the quality of clustering and (pairs of) gene ranking when using \tomatomp\ without 
explicitly specifying the $\delta$ parameter in the neighborhood graph $G_\delta$, as an illustration of Corollary~\ref{cor:metric}. More specifically, we compute \tomatomp\ clusters with a range of $\delta$ ranging from either the 1\% to the 5\% quantiles of the positive data distance distributions (datasets $(i)-(iii)$), or to ranges of lengths $2*10^{-2}$ (kpmp) and $2*10^{-1}$ (spat) around the 1\% quantiles.
We compare against the mean score of \tomato\ clusterings obtained after testing every line, as well as the worst score across lines.
This second baseline is very pessimistic, and only illustrates how poor \tomato\ clusters can be for unlucky line picks.
Moreover, we compare against three non-topological baselines: $k$-means (KM), spectral clustering (SC), and hierarchical clustering (HC). In datasets $(i)$-$(iii)$, we provide scores after running non-topological baselines both with (w) and without (w/) the function values as extra coordinates.

\begin{table}[h]
\centering
\scriptsize
\setlength{\tabcolsep}{3pt}
\caption{\label{tab:ami-metric-clus} AMI scores}
\resizebox{\textwidth}{!}{%
\begin{tabular}{|l|cccccc|cc|c|}
\hline
Dataset & KM(w/) & KM(w) & SC(w/) & SC(w) & HC(w/) & HC(w) & \tomato\ (min) & \tomato\ (mean) & \tomatomp \\ % (100) & AuToMATo (500) & AuToMATo (1000) \\
\hline
synthetic & 0.3625 & 0.6894 & 0.4499 & 0.1192 & -0.0018 & 0.1141 & 0.2351 & 0.8138 & \textbf{0.9256} \\ %& \textbf{0.9307} & 0.9273 \\
3dshape & 0.5578 & 0.5729 & 0.5598 & 0.5358 & 0.5723 & 0.5598 & 0.0341 & 0.4355 & \textbf{0.7767} \\ %& \textbf{0.7767} & \textbf{0.7767} \\
\hline
\end{tabular}%
}
\end{table}

% Auto-generated by aggregate_results_to_latex.py
\begin{table}[h]
\centering
\scriptsize
\setlength{\tabcolsep}{3pt}
\caption{\label{tab:top10-metric-rank} TopHits@10}
%\resizebox{\textwidth}{!}{%
\begin{tabular}{|l|c|cc|c|}
\hline
Dataset & HC & \tomato\ (min) & \tomato\ (mean) & \tomatomp \\ %& AuToMATo (500, mode 1) & AuToMATo (1000, mode 1) & AuToMATo (100, mode 3) & AuToMATo (500, mode 3) & AuToMATo (1000, mode 3) & AuToMATo (100, mode 4) & AuToMATo (500, mode 4) & AuToMATo (1000, mode 4) & AuToMATo (100, mode 6) & AuToMATo (500, mode 6) & AuToMATo (1000, mode 6) \\
\hline
1-g kpmp & 0.0000 & 0.8000 & \textbf{0.8800} & 0.8000 \\ %& 0.8000 & 0.8000 & 0.8000 & 0.8000 & 0.8000 & 0.8000 & 0.8000 & 0.8000 & 0.8000 & 0.8000 & 0.8000 \\
1-g spat & 0.0000 & \textbf{1.0000} & \textbf{1.0000} & \textbf{1.0000} \\ %& \textbf{1.0000} & \textbf{1.0000} & \textbf{1.0000} & \textbf{1.0000} & \textbf{1.0000} & \textbf{1.0000} & \textbf{1.0000} & \textbf{1.0000} & \textbf{1.0000} & \textbf{1.0000} & \textbf{1.0000} \\
%st1gr kpmp (200) & 0.0000 & 0.8000 & \textbf{0.8800} & 0.8000 & 0.8000 & 0.8000 & 0.8000 & 0.8000 & 0.8000 & 0.8000 & 0.8000 & 0.8000 & 0.8000 & 0.8000 & 0.8000 \\
%st1gr spatial (200) & 0.0000 & \textbf{1.0000} & \textbf{1.0000} & \textbf{1.0000} & \textbf{1.0000} & \textbf{1.0000} & \textbf{1.0000} & \textbf{1.0000} & \textbf{1.0000} & \textbf{1.0000} & \textbf{1.0000} & \textbf{1.0000} & \textbf{1.0000} & \textbf{1.0000} & \textbf{1.0000} \\
2-g kpmp & 0.0000 & 0.7000 & \textbf{0.9330} & 0.4000 \\ %& 0.1000 & 0.3000 & 0.4000 & 0.1000 & 0.3000 & 0.4000 & 0.1000 & 0.3000 & 0.4000 & 0.1000 & 0.3000 \\
2-g spat & 0.0000 & \textbf{1.0000} & \textbf{1.0000} & 0.8000 \\ %& 0.8000 & 0.8000 & 0.8000 & 0.8000 & 0.8000 & 0.8000 & 0.8000 & 0.8000 & 0.8000 & 0.8000 & 0.8000 \\
%st2gr kpmp (20) & 0.0000 & 0.8000 & \textbf{0.8710} & 0.0000 & 0.0000 & 0.0000 & 0.0000 & 0.0000 & 0.0000 & 0.0000 & 0.0000 & 0.0000 & 0.0000 & 0.0000 & 0.0000 \\
%st2gr spatial (20) & -- & -- & -- & -- & -- & -- & -- & -- & -- & -- & -- & -- & -- & -- & -- \\
\hline
\end{tabular}%
%}
\end{table}

AMI scores can be found in Table~\ref{tab:ari-metric-clus} (and ARI scores in Appendix~\ref{app:add-table}). %and~\ref{tab:scores-metric-rank}. %and running times can be found in Table~\ref{tab:times-metric}.
As one can see from Table~\ref{tab:ami-metric-clus}, \tomatomp\ can achieve favorable results against \tomato\ without any graph tuning, and against non-topological baselines (launched with default parameters using the Scikit-Learn implementations) struggle to capture the cluster structures. Importantly, concerning \tomato, a wrong choice of $\delta$ can lead to very poor clusters, and even averaging the results is not as efficient as \tomatomp, as the mean score is dragged down by those poor choices.

The TopHits@10 scores of Table~\ref{tab:top10-metric-rank} are not as good (Pearson correlations are in Appendix~\ref{app:add-table}). 
%While it is expected that the Pearson correlation between the rankings is low for pairs of genes (as our proposed multi-parameter ranking criterion might capture potentially different information than Jaccard similarities) 
The highest ranked pairs have low intersection with the ground-truth ones for kpmp, but still manage to recover $4$ pairs out of the best $10$. On the other hand, dendrogram distances computed on cell coordinates and gene expression completely fail at capturing relevant pairs.

It should be noted however, that the running times of \tomatomp\ are substantially longer: the fact that the algorithm requires barycentric subdivisions of large graphs as inputs and the cubic complexity leads to running times that are between 3 and 4 orders of magnitude longer than both non-topological and topological baselines (which run in around $10^{-1}$ seconds while \tomatomp\ takes $10^3$ seconds in the worst cases). See also Figure~\ref{fig:time-line} in Appendix~\ref{app:add-lines}.

\paragraph*{Outlier robustness}%\label{sec:exp-outliers}
In this second set of experiments, we measure the quality of clustering and ranking when using \tomatomp\ using an outlier detection score function (computed as the average absolute difference between the function value of a data point and the ones of its neighbors in $G$) after datasets have been randomly plagued with a small fraction of outlier values ($10$ for each dataset, with results averaged across 10 trials), as an illustration of Corollary~\ref{cor:outliers}.

% Auto-generated by aggregate_results_to_latex.py
\begin{table}[h]
\centering
\scriptsize
\setlength{\tabcolsep}{3pt}
\caption{\label{tab:ami-outliers-clus} AMI scores}
\resizebox{\textwidth}{!}{%
\begin{tabular}{|l|cccccc|c|c|}
\hline
Dataset & KM(w/) & KM(w) & SC(w/) & SC(w) & HC(w/) & HC(w) & \tomato & \tomatomp \\ % (100) & AuToMATo (500) & AuToMATo (1000) \\
\hline
synthetic & 0.3864 $\pm$ 0.0256 & 0.6718 $\pm$ 0.0022 & 0.4238 $\pm$ 0.0321 & 0.2234 $\pm$ 0.0140 & 0.0000 $\pm$ 0.0000 & 0.2053 $\pm$ 0.0403 & 0.5902 $\pm$ 0.1217 & \textbf{0.9268 $\pm$ 0.0117} \\ % & \textbf{0.9271 $\pm$ 0.0111} & 0.9268 $\pm$ 0.0114 \\
image & 0.4149 $\pm$ 0.0091 & 0.4015 $\pm$ 0.0118 & -- & -- & 0.4003 $\pm$ 0.0127 & 0.3860 $\pm$ 0.0171 & 0.2157 $\pm$ 0.1116 & \textbf{0.9311 $\pm$ 0.0274} \\ % & 0.9326 $\pm$ 0.0265 & \textbf{0.9332 $\pm$ 0.0258} \\
\hline
\end{tabular}%
}
\end{table}

% Auto-generated by aggregate_results_to_latex.py
\begin{table}[h]
	\centering
	\scriptsize
	\setlength{\tabcolsep}{3pt}
	\caption{\label{tab:top10-outliers-rank} TopHits@10}
	%\resizebox{\textwidth}{!}{%
		\begin{tabular}{|l|cc|c|c|}
			\hline
			Dataset & HC(w/) & HC(w) & \tomato & \tomatomp \\ %AuToMATo (100, mode 1) & AuToMATo (500, mode 1) & AuToMATo (1000, mode 1) & AuToMATo (100, mode 3) & AuToMATo (500, mode 3) & AuToMATo (1000, mode 3) & AuToMATo (100, mode 4) & AuToMATo (500, mode 4) & AuToMATo (1000, mode 4) & AuToMATo (100, mode 6) & AuToMATo (500, mode 6) & AuToMATo (1000, mode 6) \\
			\hline
			1-g kpmp & 0.0171 $\pm$ 0.0382 & 0.0000 $\pm$ 0.0000 & 0.5143 $\pm$ 0.0810 & \textbf{0.9943 $\pm$ 0.0236} \\ % & \textbf{0.9943 $\pm$ 0.0236} & \textbf{0.9943 $\pm$ 0.0236} & 0.5800 $\pm$ 0.1389 & 0.5371 $\pm$ 0.1477 & 0.5457 $\pm$ 0.1172 & 0.3371 $\pm$ 0.1395 & 0.0429 $\pm$ 0.0739 & 0.0914 $\pm$ 0.1337 & 0.3600 $\pm$ 0.0946 & 0.0486 $\pm$ 0.0781 & 0.2371 $\pm$ 0.1031 \\
			1-g spat & 0.0000 $\pm$ 0.0000 & 0.0000 $\pm$ 0.0000 & 0.8647 $\pm$ 0.0931 & \textbf{0.9765 $\pm$ 0.0437} \\ % & \textbf{0.9765 $\pm$ 0.0437} & \textbf{0.9765 $\pm$ 0.0437} & 0.8647 $\pm$ 0.0931 & 0.8647 $\pm$ 0.0931 & 0.8647 $\pm$ 0.0931 & \textbf{0.9765 $\pm$ 0.0437} & \textbf{0.9765 $\pm$ 0.0437} & \textbf{0.9765 $\pm$ 0.0437} & 0.8647 $\pm$ 0.0931 & 0.8647 $\pm$ 0.0931 & 0.8647 $\pm$ 0.0931 \\
			%st1gr kpmp (200) & 0.0125 $\pm$ 0.0354 & 0.0000 $\pm$ 0.0000 & 0.3000 $\pm$ 0.0926 & \textbf{1.0000 $\pm$ 0.0000} & \textbf{1.0000 $\pm$ 0.0000} & \textbf{1.0000 $\pm$ 0.0000} & 0.5125 $\pm$ 0.1246 & 0.4500 $\pm$ 0.0756 & 0.4250 $\pm$ 0.0707 & 0.2500 $\pm$ 0.1309 & 0.0000 $\pm$ 0.0000 & 0.0125 $\pm$ 0.0354 & 0.1125 $\pm$ 0.0354 & 0.0125 $\pm$ 0.0354 & 0.1750 $\pm$ 0.0463 \\
			%st1gr spatial (200) & 0.0000 $\pm$ 0.0000 & 0.0000 $\pm$ 0.0000 & 0.8688 $\pm$ 0.0946 & \textbf{0.9750 $\pm$ 0.0447} & \textbf{0.9750 $\pm$ 0.0447} & \textbf{0.9750 $\pm$ 0.0447} & 0.8688 $\pm$ 0.0946 & 0.8688 $\pm$ 0.0946 & 0.8688 $\pm$ 0.0946 & \textbf{0.9750 $\pm$ 0.0447} & \textbf{0.9750 $\pm$ 0.0447} & \textbf{0.9750 $\pm$ 0.0447} & 0.8688 $\pm$ 0.0946 & 0.8688 $\pm$ 0.0946 & 0.8688 $\pm$ 0.0946 \\
			2-g kpmp & 0.0588 $\pm$ 0.0507 & 0.0000 $\pm$ 0.0000 & 0.1824 $\pm$ 0.0809 & \textbf{0.5059 $\pm$ 0.0556} \\ % & 0.5000 $\pm$ 0.0707 & 0.5000 $\pm$ 0.0866 & 0.2000 $\pm$ 0.0000 & 0.1882 $\pm$ 0.0600 & 0.1882 $\pm$ 0.0600 & 0.2118 $\pm$ 0.1054 & 0.1000 $\pm$ 0.0866 & 0.1059 $\pm$ 0.0899 & 0.1235 $\pm$ 0.0831 & 0.0941 $\pm$ 0.0899 & 0.2059 $\pm$ 0.0429 \\
			2-g spat & 0.0000 $\pm$ 0.0000 & 0.0000 $\pm$ 0.0000 & 0.7875 $\pm$ 0.0991 & \textbf{0.8125 $\pm$ 0.0354} \\ % & \textbf{0.8125 $\pm$ 0.0354} & \textbf{0.8125 $\pm$ 0.0354} & \textbf{0.8125 $\pm$ 0.2167} & 0.2125 $\pm$ 0.2850 & 0.3750 $\pm$ 0.3196 & 0.8000 $\pm$ 0.0535 & 0.8000 $\pm$ 0.0535 & \textbf{0.8125 $\pm$ 0.0641} & 0.7625 $\pm$ 0.1685 & 0.2750 $\pm$ 0.3240 & 0.4375 $\pm$ 0.3204 \\
			%st2gr kpmp (20) & 0.0000 $\pm$ 0.0000 & 0.0000 $\pm$ 0.0000 & 0.1000 $\pm$ 0.0000 & \textbf{0.3333 $\pm$ 0.1155} & \textbf{0.3333 $\pm$ 0.1155} & \textbf{0.3333 $\pm$ 0.0577} & 0.1000 $\pm$ 0.0000 & 0.0667 $\pm$ 0.0577 & 0.0667 $\pm$ 0.0577 & 0.0000 $\pm$ 0.0000 & 0.0000 $\pm$ 0.0000 & 0.0000 $\pm$ 0.0000 & 0.0000 $\pm$ 0.0000 & 0.0000 $\pm$ 0.0000 & 0.0667 $\pm$ 0.0577 \\
			%st2gr spatial (20) & 0.0000 $\pm$ 0.0000 & 0.0000 $\pm$ 0.0000 & \textbf{0.6000 $\pm$ 0.0000} & 0.2000 $\pm$ 0.0000 & 0.2000 $\pm$ 0.0000 & 0.2000 $\pm$ 0.0000 & 0.2500 $\pm$ 0.0707 & 0.0500 $\pm$ 0.0707 & 0.4000 $\pm$ 0.1414 & 0.2000 $\pm$ 0.0000 & 0.2000 $\pm$ 0.0000 & 0.2500 $\pm$ 0.0707 & 0.3000 $\pm$ 0.1414 & 0.0500 $\pm$ 0.0707 & 0.3500 $\pm$ 0.2121 \\
			\hline
		\end{tabular}%
		%}
\end{table}

AMI scores can be found in Table~\ref{tab:ami-outliers-clus} (and ARI scores in Appendix~\ref{app:add-table}). 
%and running times can be found in Table~\ref{tab:times-outlier}.
As one can see from both tables, non-topological and topological baselines definitely struggle in the face of outlier values (spectral clustering did not terminate after a few hours of computations), because outlier values are immediately detected as persistent modes (see also Figure~\ref{fig:outlier-results} in Appendix~\ref{app:add-plot}). On the other hand, \tomatomp\ clusterings remain highly accurate.
As for rankings, Pearson correlation of \tomatomp\ is again low for gene pairs but top hits remain competitive: while the rankings of \tomato\ drop to less than $2$ detected relevant pairs out of $10$ in average for kpmp, \tomatomp\ rankings stay (in average) above $5$, despite being lowly correlated to the ground-truth CoSS.

As in the previous section, running times are still much longer for \tomatomp, but as barycentric subdivisions are not required anymore, the difference is reduced to $10^{-1}$ seconds for baselines against $10^2$ seconds for \tomatomp.  See also Figure~\ref{fig:time-line} in Appendix~\ref{app:add-lines}.

\begin{figure}[h!]
	\centering
	\includegraphics[width=6cm]{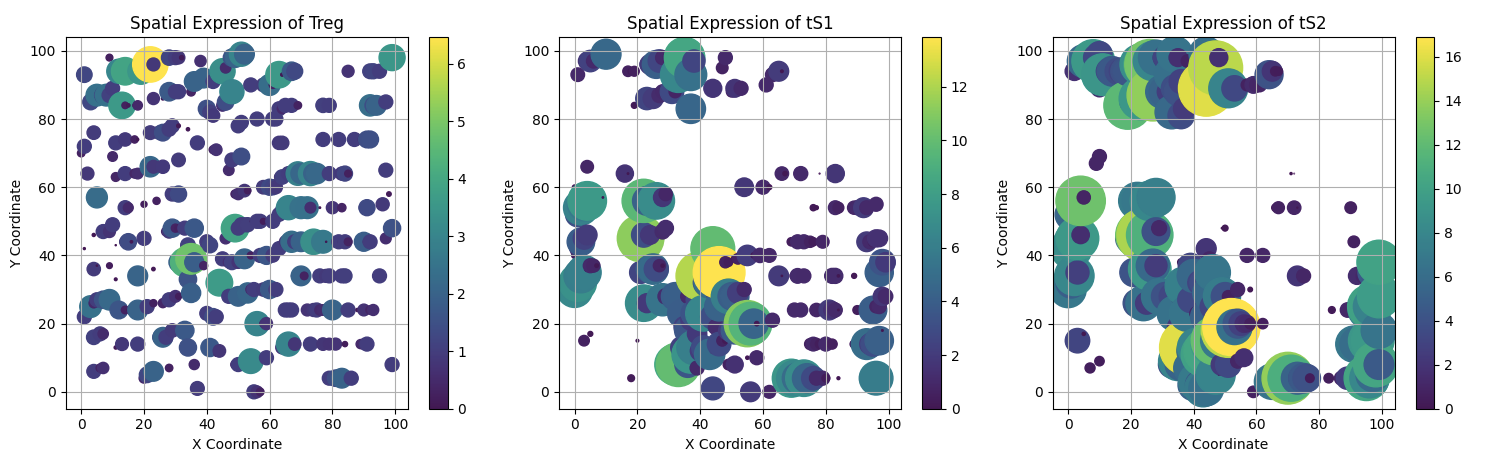}\ \ \ \ \ \ \ 
	\includegraphics[width=6cm]{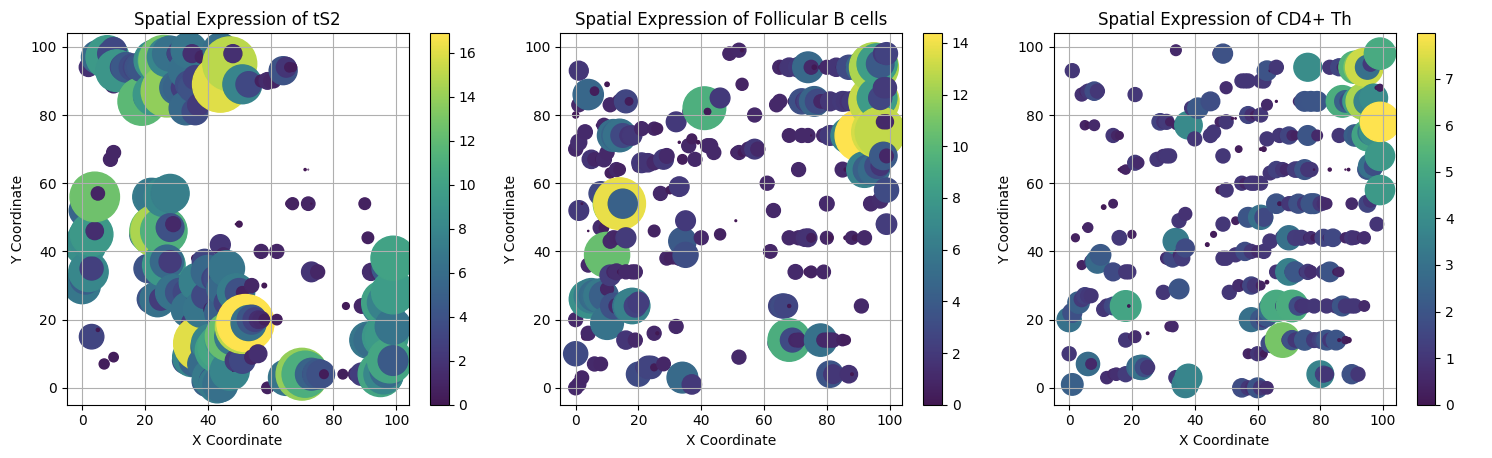}
	\caption{{\textbf{(Left)}} Highly ranked triplet of genes of the spat dataset. All three genes co-localize on the top-left and middle-down tissue regions. {\textbf{(Right)}} Lower ranked triplet.} %Whenever one gene is highly expressed in a region, at least one of the other two is not expressed in that region.}
	\label{fig:triplets}
\end{figure}

\paragraph*{Multiple functions.}%\label{sec:exp-tomatomp}
Finally, as a qualitative illustration of the high flexibility of \tomatomp, we end this numerical experiment section by performing rankings of \emph{triplets of genes}, using $3$ different gene expression functions at the same time. One can see in Figure~\ref{fig:triplets} that for spat, the top ranked triplets do characterize genes that co-localize in the tissue, while lower ranked triplets have at least one gene in the triplet that is not highly expressed where the other two are. See Figure~\ref{fig:triplets-kpmp} in Appendix~\ref{app:add-plot} for a similar example with kpmp.

%focus on spatial transcriptomics datasets, and showcase how $2$-parameter \tomatomp\ using two marker genes compare to using Jaccard similarities (as presented at the beginning of the section). We also provide the first (to our knownledge) rankings for gene triplets using 3-parameter \tomatomp.

%\input{latex_aggregated_by_experiment_method}
%\input{latex_aggregated_by_experiment_method_tables/table_no_radius_tophits_20}
%\input{latex_aggregated_by_experiment_method_tables/table_no_radius_tophits_30}
%\input{latex_aggregated_by_experiment_method_tables/table_outliers_tophits_20}
%\input{latex_aggregated_by_experiment_method_tables/table_outliers_tophits_30}

%--------------------------------------------------
\section{Conclusion}~\label{sec:conclusion}

In this article, we introduced \tomatomp, the first topological clustering method able to handle several functions at the same time. We showed that, similar to the standard \tomato\ algorithm, \tomatomp\ enjoys robustness guarantees, and that it compares favorably to both topological and non-topological baselines on various datasets. Several directions remain open for future work:

\begin{itemize}
	\item The long running times of \tomatomp\ clearly leave room for improvement. We believe that leveraging recent, fast optimizations of persistent homology computations could help reducing this limitation~\cite{deyComputingZigzagVineyard2024}.
	\item Relaxing assumption $(A1)$ of Theorem~\ref{thm:tomatomp} would be a considerable theoretical improvement, as this assumption is going to be harder to satisfy as the number of functions increases. We believe this could be achieved upon careful study of the matching functions $\{\matching_\ell\}_{\ell\in\linefam}$.
	\item The study of co-localizing gene tuples is promising and will be further developed in future work, as we believe that \tomatomp\ might help for the joint analysis of multiple genes without going through expensive pairwise comparisons. Applications to tissue segmentation is also another potential avenue for \tomatomp\ in spatial transcriptomics.
\end{itemize}

%MP-ToMATo extends persistence-based clustering to multiparameter settings with theoretical guarantees and practical advantages.

\bibliographystyle{plain}
\bibliography{biblio}

\newpage
\appendix

\section{Details on \tomato\ and (multi-parameter) PH}\label{app:background}

\paragraph{Persistence-Based Clustering.} %\label{sec:tomato}
Topological clustering, and its corresponding algorithm, \textsc{ToMATo}\ (Topological Mode Analysis Tool) was introduced in~\cite{chazalPersistenceBasedClusteringRiemannian2013} as a clustering method blending hierarchical clustering and mode-seeking algorithms. The core idea of \tomato\ is to compute a hierarchy on the modes, or basins of attraction, of a given density estimator $f:\Xn\to\R$ (or, more generally, of any continuous function). Moreover, instead of a dendrogram (as in traditional hierarchical clustering), this hierarchy is encoded in a so-called \emph{persistence diagram} $\dgm_0(f)$.

%\paragraph{The \tomato\ algorithm.} 
Roughly speaking, \tomato\ implements the following steps.

First, a graph $G$ is computed on the data points, whose vertex set $V(G)$ is the dataset $\Xn$ itself. This graph is most of the time computed as a $\delta$-neighborhood graph for some $\delta > 0$, but any user-provided graph can be used.

Then, \tomato\ orders the data points based on their function values, from highest to lowest, and processes them one by one. The first ones (resp. last ones) to be considered are thus the local maxima (resp. minima) of the function. This sequence of growing subsets of the dataset is often called a \emph{filtration} of the dataset.

Now, given some current point $x$, the graph $G$ is first used to decide the type of $x$.
\begin{itemize}
	\item If the function values of the neighbors of $x$ are all below (resp. above) $f(x)$, its type is set as local maxima (resp. minima), and a new mode is introduced.
	\item If the set of neighbors have both larger and lower values, then the modes of the neighbors with larger function values, and the local maxima that they are associated with, are identified\footnote{This is typically done using a \textsc{Union-Find} data structure}. If only one mode (resp. several modes) is found, the type of $x$ is set as regular point (resp. saddle).
\end{itemize}

Finally, if $x$ is a saddle, all modes are merged to the mode whose corresponding local maximum has highest function value. See Algorithm~\ref{alg:tomato} in Appendix~\ref{app:background} for some pseudo-code.
%by making the local maxima point to the mode with highest local maximum they are merged with. 

As one can see, \tomato\ creates a hierarchy between the modes, as every mode (except the one associated to the global maximum) is pointing to the one that it is eventually merged with. Moreover, one can record, for each mode: $(a)$ the function value of its local maximum, called its \emph{birth time}, and $(b)$ the function value of its saddle point inducing its merge, called its \emph{death time} (if the mode is not associated to the global maximum). The \emph{persistence diagram} (PD) of $f$, denoted as $\dgm_0(f)$ records birth and death times with points in $\R^2$: every PD point represents a mode, whose birth and death times are used as coordinates (the PD points are thus below the diagonal, as death time is always lower than birth time). The absolute difference between the coordinates of a PD point $p$ is usually called the \emph{prominence} $\prom(p)$ of the corresponding mode. 

Finally, a clustering of $\Xn$ can be obtained 
%either by specifying a number of clusters, or 
by setting a threshold $\tau$ on the PD point prominences. In this case, the PD points with prominences at least $\tau$ are selected, and the corresponding modes are used as clusters. Then, for every remaining PD point and associated modes, the hierarchy built by \tomato\ is used to decide the PD point and corresponding cluster (with prominence at least $\tau$) that it should be merged with. 
%See Figure~\ref{fig:tomato} (and Algorithm~\ref{alg:tomato} in Appendix~\ref{app:background}).
Note that a nice feature of \tomato\ is its complexity: using a \textsc{Union-Find} data structure, computing the hierarchy of \tomato\ can be done in quasilinear time. 

\begin{algorithm}
	\caption{The \tomato\ algorithm}\label{alg:tomato}
	\KwData{$\Xn, f, G, \tau$}
	\KwResult{$\clustering_{\tomato}$}
	Sort $\{x_1,\dots,x_n\}$ in decreasing order of $f$: $f(x_1)\geq\dots\geq f(x_n)$\;
	Initialize a union-find data structure $\mathcal{U}$ and two lists $g,r$ of length $n$\;
	\For{$1\leq i\leq n$}{
		$\mathcal N\leftarrow$ neighbors of $x_i$ in $G$ that have indices lower than $i$\;
		\If{$\mathcal N=\varnothing$}{
			Create a new entry $e$ in $\mathcal{U}$ and attach vertex $x_i$ to it: ${\mathcal U}$.{\tt MakeSet}$(x_i)$\;
			$r[e]\leftarrow x_i$\;
		}
		\Else{
			$g[i] \leftarrow {\rm argmax} \{f(x_j):x_j\in \mathcal{N}\}$\;
			$e_i \leftarrow \mathcal{U}$.{\tt Find}$(g[i])$\;
			Attach vertex $x_i$ to the entry $e_i$: ${\mathcal U}$.{\tt Union}$(x_i,e_i)$\;
			\For{$x_j\in\mathcal{N}$}{
				$e \leftarrow \mathcal{U}$.{\tt Find}$(x_j)$\;
				\If{$e\neq e_i$ \textbf{and} $\min\{f(r[e]),\; f(r[e_i])\} < f(x_i) + \tau$}{
					$\mathcal{U}$.{\tt Union}$(e,\;e_i)$\;
					$r[e\cup e_i] \leftarrow {\rm argmax} \{f(r[e]),\; f(r[e_i])\}$\;
					$e_i \leftarrow e\cup e_i$
				}
			}
		}
	}
	\For{$e\in\mathcal U$}{
		\For{$x\in e$}{
			$\clustering_{\tomato}(x)\leftarrow r[e]$\;
		}
	}
\end{algorithm}

\paragraph{Diagram distances.} Diagram distances are defined by finding partial correspondences between PD points, and picking the one with lowest cost.

\begin{definition}[$q$-th diagram distance]\label{def:pd-dist}
	Let $\dgm$ and $\dgm'$ be two PDs. The \emph{$q$-th diagram distance} $d_q$ between PDs is defined as:
	\begin{equation}
		d_q(\dgm,\dgm'):=\inf_{P\in\mathcal{P}(\dgm,\dgm')} c_q(P),
	\end{equation} 
	where $\mathcal{P}(\dgm,\dgm')$ denotes the set of \emph{partial correspondences} between $\dgm$ and $\dgm'$, i.e., the set of subsets of $\dgm\times\dgm'$ s.t. the first and second projections $\pi_1:(p,p')\mapsto p$ and $\pi_2:(p,p')\mapsto p'$ are injective, and where the \emph{$q$-th cost} of a partial correspondence $P$ is defined as:
	\begin{equation}\label{eq:cost}
		c_q(P)^q:=\sum_{(p,p')\in P}\|p-p'\|_\infty^q + 
		\sum_{p\not\in\im(\pi_1)} \|p-\pi_\Delta(p)\|_\infty^q +
		\sum_{p'\not\in\im(\pi_2)} \|p'-\pi_\Delta(p')\|_\infty^q.
	\end{equation}
	When $q\to+\infty$, the $q$-th diagram distance is called the \emph{bottleneck distance} $d_b$ between PDs (and can be computed using maximum instead of sum in Equation~(\ref{eq:cost})).
\end{definition}

\section{Proofs of Section~\ref{sec:theory}}\label{app:proofs}

The proof of Theorem~\ref{thm:tomatomp} follows from a few results on \tomato\ and multi-parameter PH from the recent TDA literature, that we now restate using our notations.

\paragraph{Robustness of \tomato.} If $\Xn\subset X$ has been sampled from $X$, and $\hat f:=f|_{\Xn}$ is the restriction of $f$ to $\Xn$, the following result connects 
%one might thus wonder whether 
the clusters of \tomato\ computed on $\Xn$ with those computed on $X$.

\begin{theorem}[{\cite[Theorem 3.2 $(iii)$]{andreEstimatingPersistentHomology2025}}]\label{thm:single-sfa}
	Let $X$ be a compact geodesic space with convexity radius $\rho(X)$, $f:X\to\R$ be a $c$-Lipschitz function, and assume $\Xn$ is a finite geodesic $\eps$-sample of $X$, i.e., every $x\in X$ is at geodesic distance at most $\eps$ from a point in $\Xn$.
	Let $\delta\in[4\eps,\rho(X)[$, and let $G_\delta$ be the corresponding $\delta$-neighborhood graph. Then,
	\begin{equation}
		d_b(\dgm_0(f), \dgm_0(\hat f))\leq c\delta.
	\end{equation}
	Moreover, if $\dgm_0(f)$ is $(d_1,d_2)$-separated and $\delta < \frac{d_2-d_1}{16c}$,
	then 
	%$\dgm_0(\hat f)$ is $(d_1+2c\delta, d_2-2c\delta)$-separated, and 
	$\tomato(\Xn,\hat f, G_\delta, \tau)$ is $3c\delta$-related to $\tomato(X,f,\tau)$, for any $d_1 < \tau < d_2$.
\end{theorem}

\paragraph{Line stability of multi-parameter PH.} A classical result of multi-parameter PH ensures that functions obtained from closed lines have close PDs.

\begin{theorem}[{\cite[Lemma 2]{landiRankInvariantStability2018}}]\label{thm:line-stab}
	Let $X$ be a topological space.
	Let $f_1,\dots,f_p:X\to\R$ be continuous functions. 
	Let $\ell, \ell'$ be two diagonal lines at distance $\eta$ from each other. Then,
	\begin{equation}\label{eq:stab-line}
		d_b(\dgm_0(\slicedfunc{\ell}{f_1,\dots,f_p}), \slicedfunc{\ell'}{f_1,\dots,f_p})) \leq\eta.
	\end{equation}
	Moreover, if $\dgm_0(\slicedfunc{\ell}{f_1,\dots,f_p})$ is $(d_1,d_2)$-separated and $\eta < (d_2-d_1)/16$, then %$\dgm_0(\slicedfunc{\ell'}{f_1,\dots,f_p})$ is $(d_1+2\eta, d_2-2\eta)$-separated, and 
	$\tomato(X,\slicedfunc{\ell}{f_1,\dots,f_p},\tau)$ and $\tomato(X,\slicedfunc{\ell'}{f_1,\dots,f_p},\tau)$ are mutually $3\eta$-related, for any $d_1 <\tau < d_2$. 
\end{theorem}

\paragraph{Proof of Theorem~\ref{thm:tomatomp}.}

%\begin{proof}
We first prove $(i)$. %show that the matching functions are unique. 

Let $\ell_i\in\linefam$. 
Theorem~\ref{thm:line-stab}, Equation~(\ref{eq:stab-line}) ensures that the points of $\dgm_0(\slicedfunc{\ell_{i+1}}{f_1,\dots,f_p})$ are included in $\|\cdot\|_\infty$-balls of radius $\eta$ around the points of 
$\dgm_0(\slicedfunc{\ell_{i}}{f_1,\dots,f_p})$.
Then, Assumption (A2) ensures that, for PD points with prominences at least $d_2$, these balls do not intersect, making the possibilities of finding matching functions with cost less than $\eta$ reduced to one candidate (the one that matches the points in $\dgm_0(\slicedfunc{\ell_{i}}{f_1,\dots,f_p})$ to the ones of $\dgm_0(\slicedfunc{\ell_{i+1}}{f_1,\dots,f_p})$ in their corresponding balls). Hence, the matching functions are unique when restricted to PD points with prominences at least $d_2$. Now these restricted matching functions must be bijections, otherwise the cost of such matching functions would exceed $\frac{d_2-d_1}{2}>\eta$.

We now prove $(ii)$ and $(iii)$.

As the matching functions are unique and thus induced by inclusions, they must coincide with the bijections introduced in
Theorem~\ref{thm:stab-tomato-clusters}. Now, let $x\in\bigcap_{\ell\in\linefam}
{\rm core}_{\ell,\tomato}(i)$, for some $1\leq i\leq N$.
Let $\ell_1$ be the first line in $\linefam$. By definition,
one has $\clustering_{\ell_1,\mma}(x) = i$.
As $x\in{\rm core}_{\ell_1,\tomato}(i)$, let $p_{\pi_{\ell_1}(i)}$ be the corresponding point of $\dgm_0(\slicedfunc{\ell_1}{f_1,\dots,f_p})$ associated to the mode that $x$ belongs to. By definition, $p_{\pi_{\ell_1}(i)}$ must have prominence at least $d_2$. Then, Theorem~\ref{thm:stab-tomato-clusters} ensures that $x$ also belongs to the cluster of $\clustering_{\ell_2,\tomato}$ associated to the point $p_j:=\matching_{\ell_1}(p_{\pi_{\ell_1}(i)})\in\dgm(\slicedfunc{\ell_2}{f_1,\dots,f_p})$,
where $1\leq j\leq N$.
Moreover, by definition, one has $j=\pi_{\ell_2}(i)$, and $\clustering_{\ell_2,\mma}(x) = i$.
%(again, $b_2$ must have prominence $d_2$). 
%Thus, $x\in \clustering_{\ell_2,\tomato}^{-1}(\pi_{\ell_2}(i))$.  and let $C_{\pi(i)}=\clustering_{\ell_1,\tomato}^{-1}(\pi_{\ell_1}(i))$
As one can apply the same reasoning iteratively to the remaining lines of $\linefam$, it follows that $\clustering_{\ell,\mma}(x) = i$ for all $\ell \in\linefam$, and thus
we finally have $\clustering_{\tomatomp}(x)=i$. 
%\end{proof}

\paragraph{Extensions of Theorem~\ref{thm:tomatomp}.}
%\begin{remark}
Theorem~\ref{thm:tomatomp} can be straightforwardly strengthened in three ways:
\begin{enumerate}
	\item[$(i)$] The result still holds even if the PDs are $(d_1,d_2)$-separated only for a \emph{subset} $\tilde\linefam$ of size at least $\frac{|\linefam|}{2}$ of consecutive lines in $\linefam$, as the most frequent cluster (computed with the argmax in Algorithm~\ref{alg:tomatomp}) for $\linefam$ will coincide with the cluster obtained for $\tilde\linefam$ with Theorem~\ref{thm:tomatomp}.
	\item[$(ii)$] If the functions $f_1,\dots,f_p$ are not provided, and one only has access to perturbed versions $\tilde f_1,\dots,\tilde f_p$ instead,
	with $\|f_i-\tilde f_i\|_\infty\leq\eps$ and $\eps\leq\frac{d_2-d_1}{4}$, then the PDs will remain $(d_1+2\eps,d_2-2\eps)$-separated, ensuring that the clusters will remain correct upon decreasing $\eta$.
	\item[$(iii)$] Concerning Assumption (A2), it might be impossible to find $\eta$ s.t. $ \eta \leq d^*$,
	%\frac 12 \min\{\|p-p'\|_\infty\,:\,p,p'\in \dgm_0(\ell,f_1,\dots,f_p), \text{ for some }\ell\in\linefam\text{ and } p,p'\text{ have prominence at least }d_2\}$, 
	as the right hand size might decrease to $0$ as $\eta$ decreases. In that case, instead of using 
	%defining the matching functions $\{\matching_\ell\}$ with 
	the $q$-th diagram distances, one can define matching functions induced by inclusions (which can be obtained using, e.g., vineyard matching functions~\cite{cohen-steinerVinesVineyardsUpdating2006} and representative cycle updates~\cite{giuntiPruningVineyardsUpdatingMonMar2300:00:00UTC2026}). %As these are induced by inclusions, the proof of the result still holds in that case.
\end{enumerate}
%\end{remark}

\begin{figure}[h!]
	\centering
	\includegraphics[width=8cm]{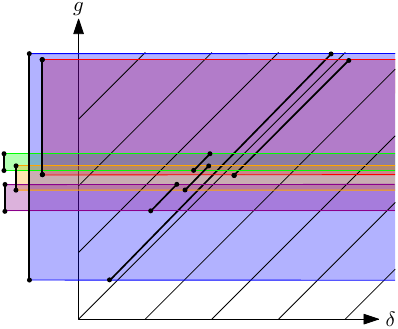}
	\caption{Sliced PDs induced by~\cite[Theorem 3.6 $(iii)$]{andreEstimatingPersistentHomology2025}}
	\label{fig:lanmodule}
\end{figure}

\paragraph{Proof of Corollary~\ref{cor:metric}.}
This corollary follows directly from the fact that, under the assumptions of Corollary~\ref{cor:metric}, the induced PDs are all $(d_1+2c\delta_{\rm max}, d_2-2c\delta_{\rm max})$-separated. This follows from the structure of the 2-parameter persistence module in degree 0 of $f$ and $\delta$, which looks like the one in Figure~\ref{fig:lanmodule}, where the PD of $g$ is at bottleneck distance $2c\delta_{\rm max}$ from the one of $f$, and is plotted with black bars on the left. As the module is obtained by taking bands stemming from the PD of $g$, induced PDs are all well-separated if the one of $f$ is, and Theorem~\ref{thm:tomatomp} applies.

\begin{figure}[h!]
	\centering
	\includegraphics[width=8cm]{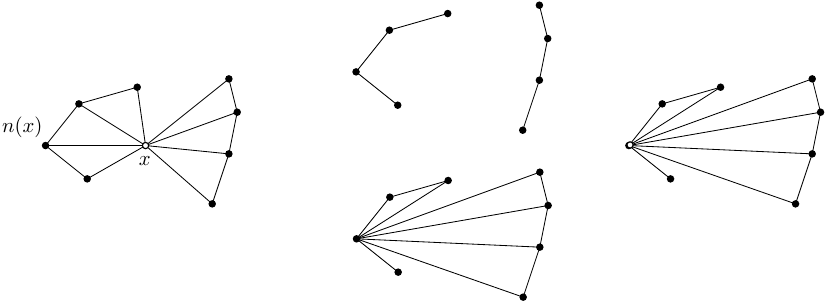} 
	\caption{\textbf{(Left)} Example of point with outlier value (empty disk) who is not topologically robust. \textbf{(Middle top)} Removing the point changes the graph topology. \textbf{(Middle bottom)} Adding edges to the leftmost neighbor makes the point topologically robust. \textbf{(Right)} Duplicating the left neighbor keeps topology unchanged.}
	\label{fig:neighbors}
\end{figure}

\paragraph{Proof of Corollary~\ref{cor:outliers}.}
This corollary follows directly from the fact that topologically robust points can be removed from $G_\delta$ without substantially changing its PDs, if the function $f$ is $c$-Lipschitz. Indeed, if a topologically robust point $x$ is removed, the resulting graph can still be turned into a graph isomorphic to $G_\delta$ by duplicating the neighbor $n(x)$, with no change in the connected components as this amounts to apply \emph{coning} on the neighborhood of $x$.
%homology groups in degree $0$. 
Letting $\tilde f$ be the corresponding function, $f$ and $\tilde f$ will be equal everywhere except on $n(x)$, on which their difference will be at most $c\delta$. Also, adding point $x$ again later in the filtration (after all other points have been processed) induce no change in the connected components as well for the same reason.
%homology groups in degree $0$. 
Hence, $d_b(\dgm_0(f),\dgm_0(\tilde f))\leq c\delta$, and combining this with the approximation factor induced by sampling from Theorem~\ref{thm:single-sfa} leads to sliced PDs similar to those in Figure~\ref{fig:lanmodule}. Thus, Theorem~\ref{thm:tomatomp} applies. See also Figure~\ref{fig:neighbors}.
%Note that, on the other hand, getting such a control over the PDs is impossible if outlier points are not topologically robust.
%Moreover, Theorem~\ref{thm:single-sfa} %3.2 $(iii)$ in \cite{andreEstimatingPersistentHomology2025} 
%shows that the PD of $f$ computed on $G_\delta$ approximates the one of $f$ computed on $X$ (if $\delta$ is appropriately chosen) with quality $c\delta$. 

%Combining the two previous facts leads to the result.

\clearpage
\section{Additional Results}\label{app:results}

\subsection{Scores and running times vs Number of lines for \tomatomp}\label{app:add-lines}

\begin{figure}[h!]
	\centering
	\includegraphics[width=6cm]{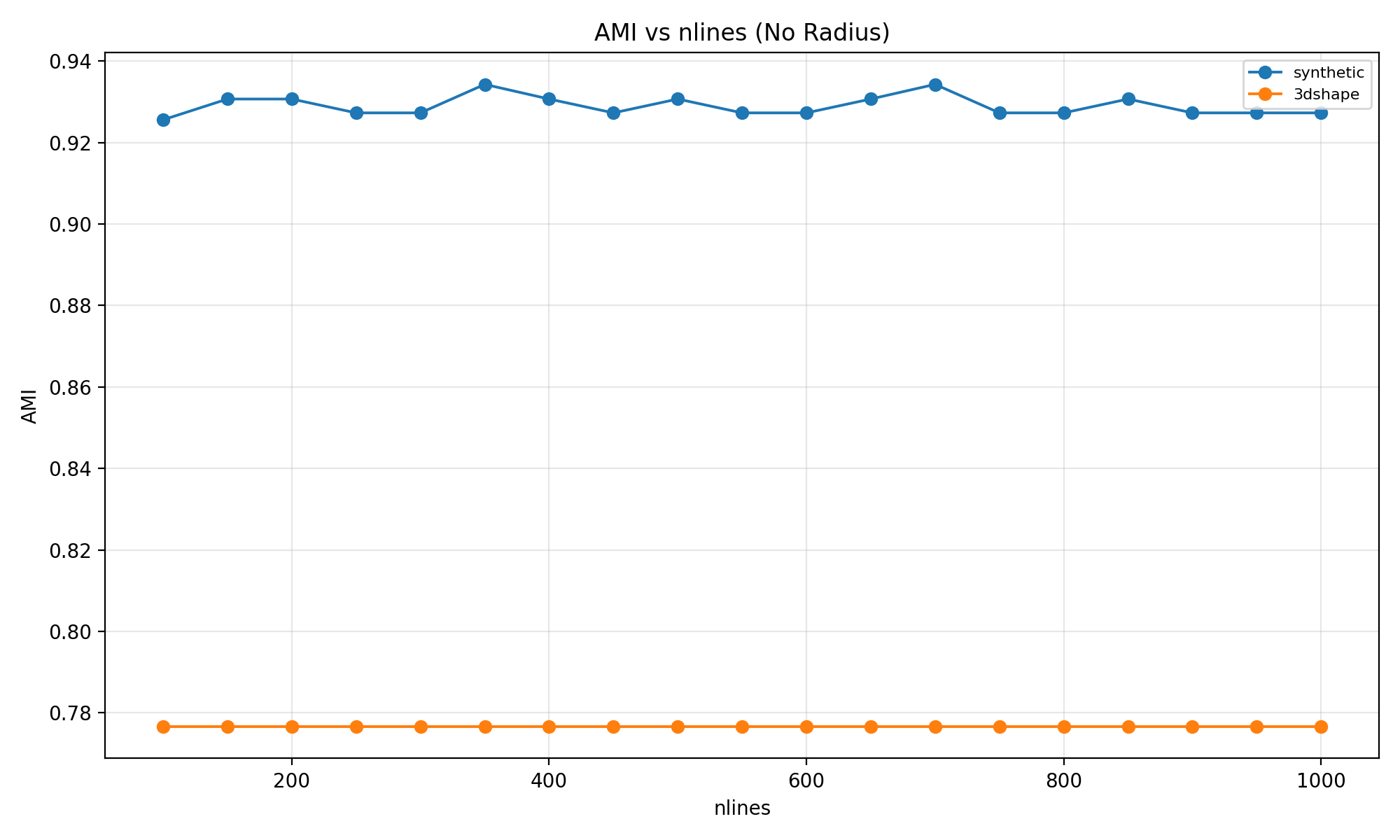} 
	\includegraphics[width=6cm]{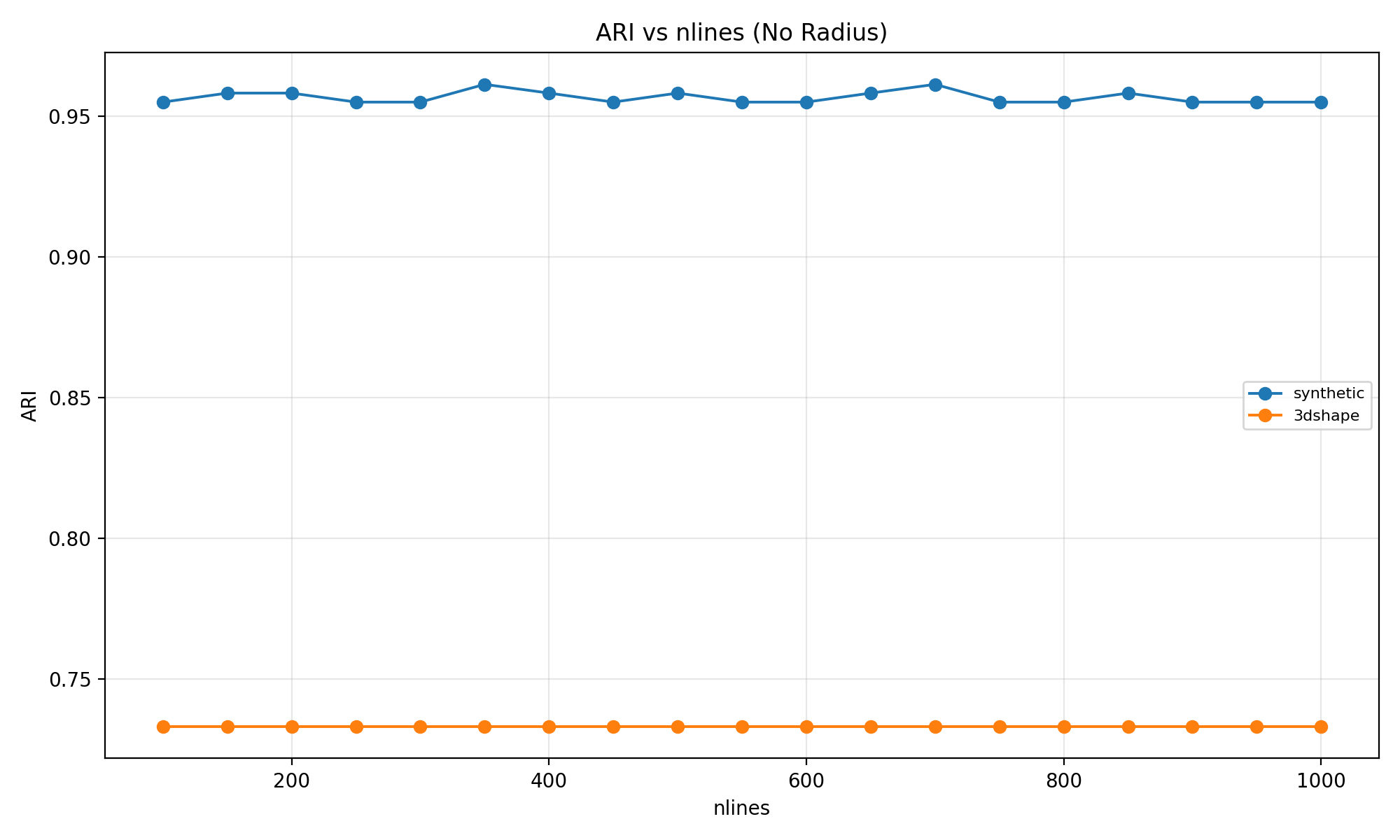} 
	\includegraphics[width=6cm]{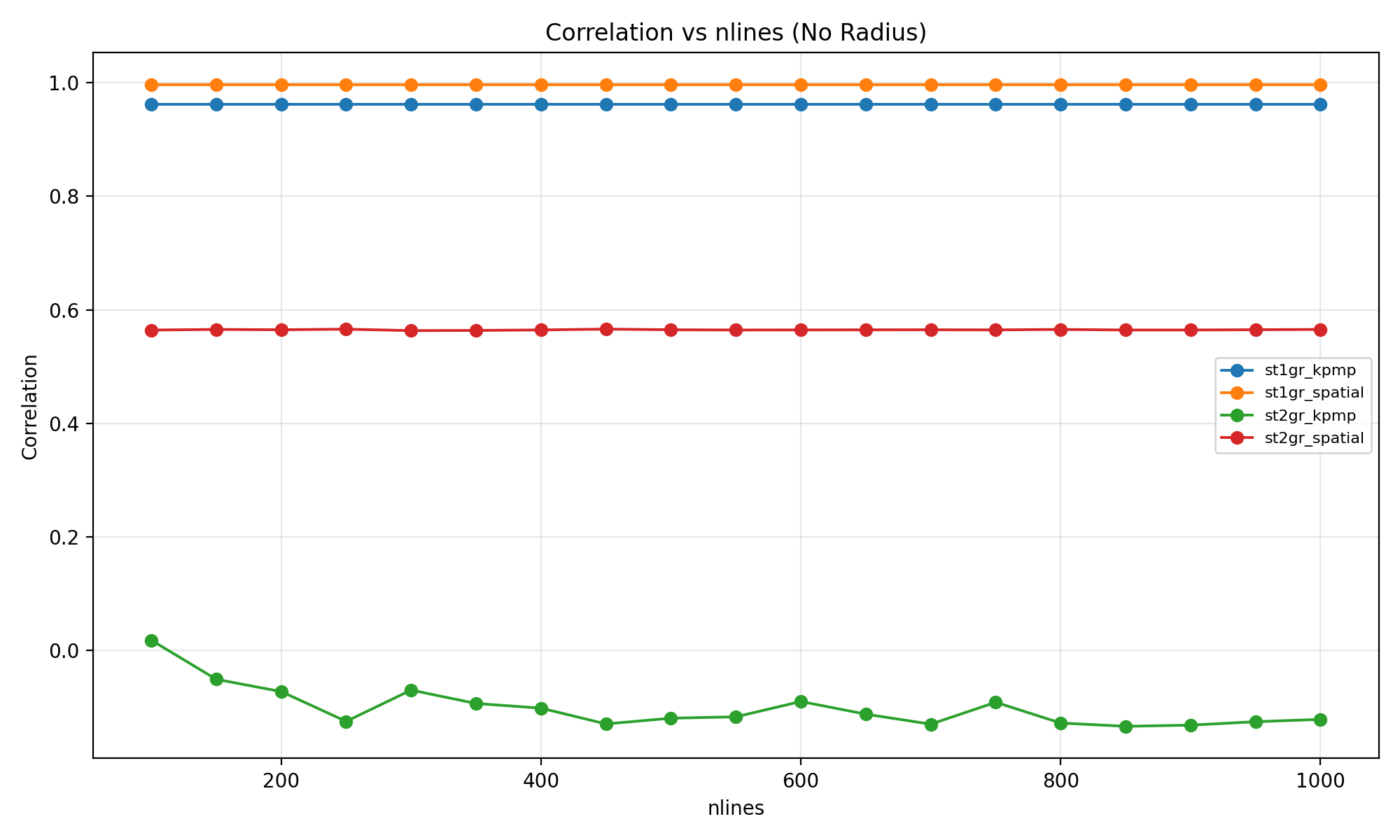} 
	\includegraphics[width=6cm]{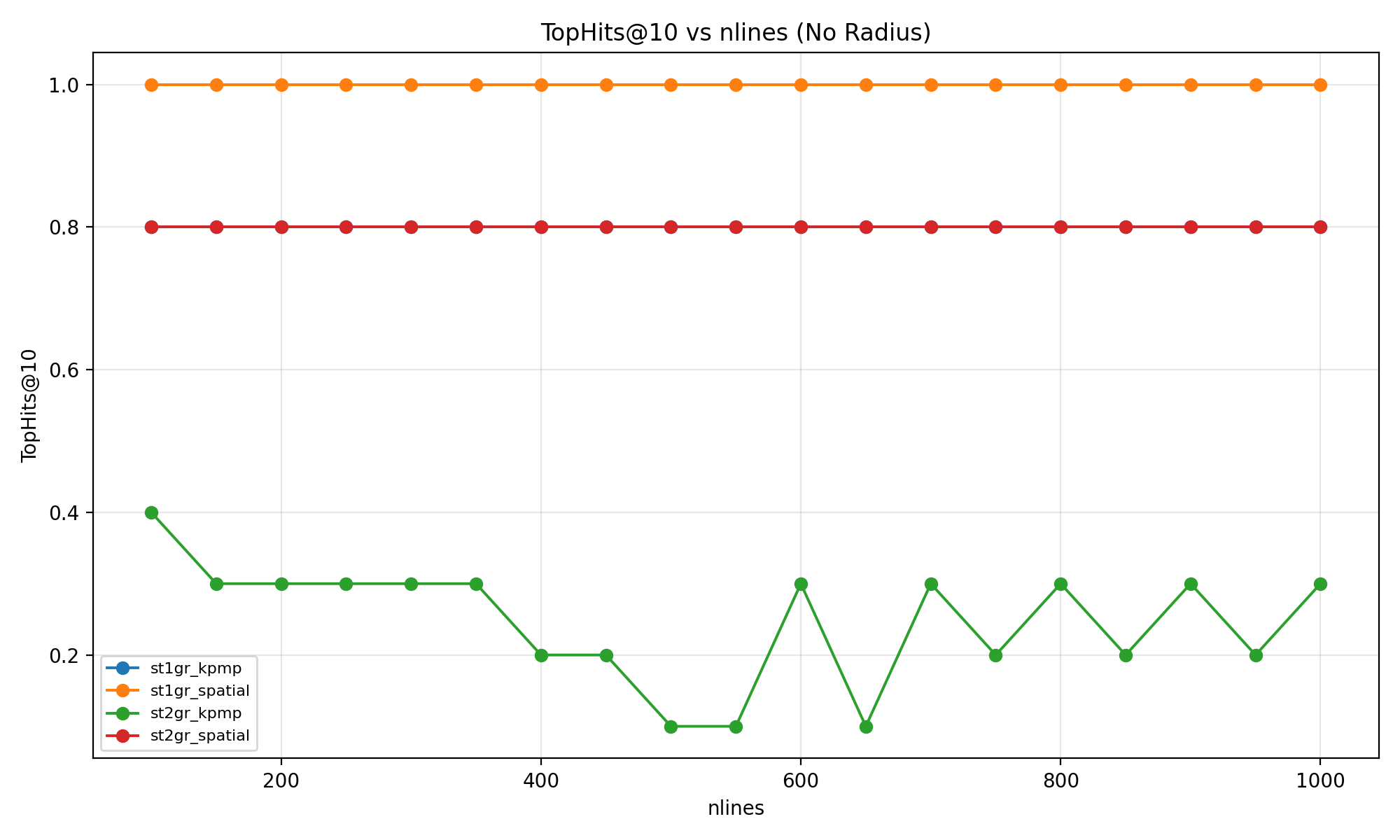} 
	\caption{Quality scores vs number of lines $|\linefam|$ for numerical experiments with no graph parameter tuning. As scores are roughly constant (except for 2-g kpmp), this motivates keeping a moderate number of lines.}
	\label{fig:score-line-no-radius}
\end{figure}

\begin{figure}[h!]
	\centering
	\includegraphics[width=6cm]{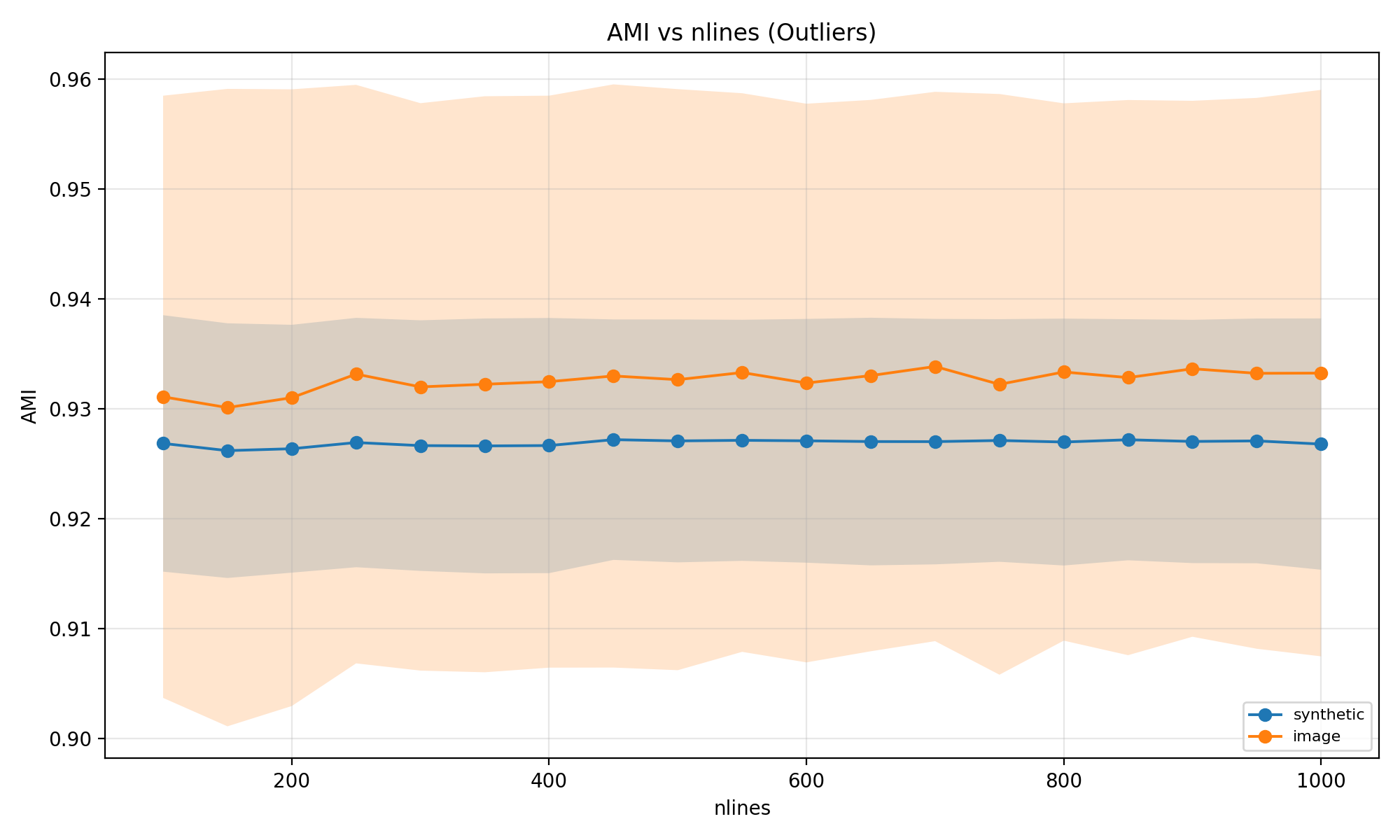} 
	\includegraphics[width=6cm]{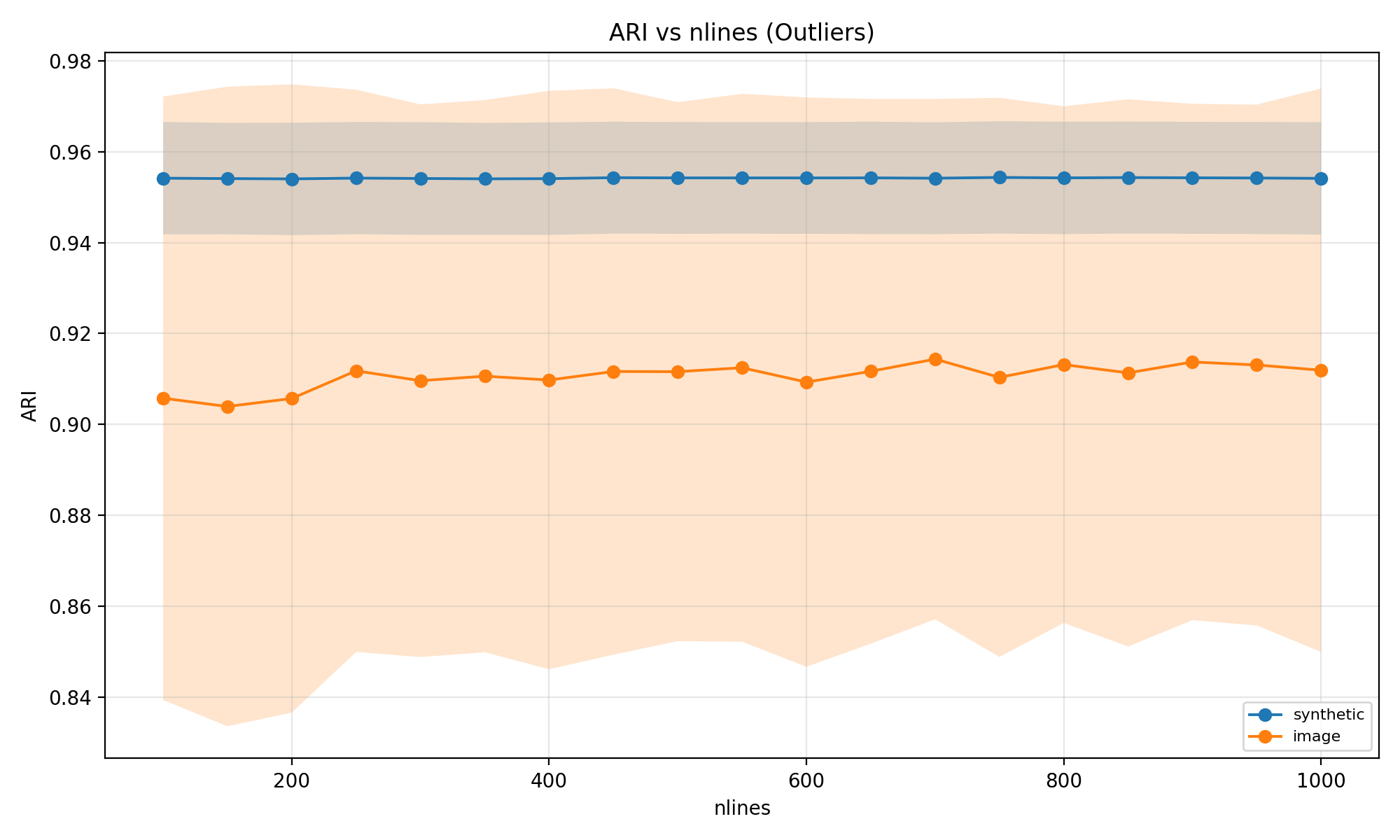} 
	\includegraphics[width=6cm]{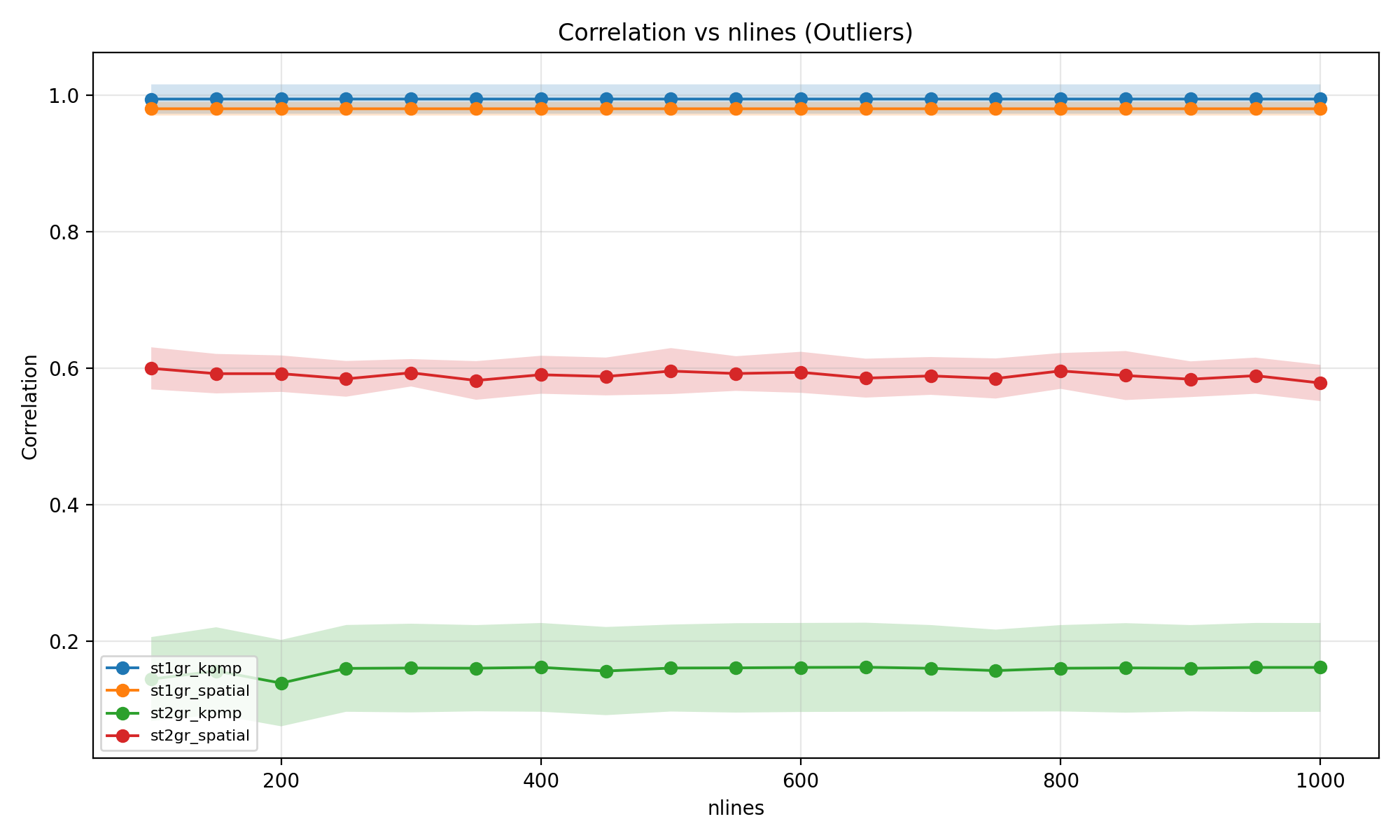} 
	\includegraphics[width=6cm]{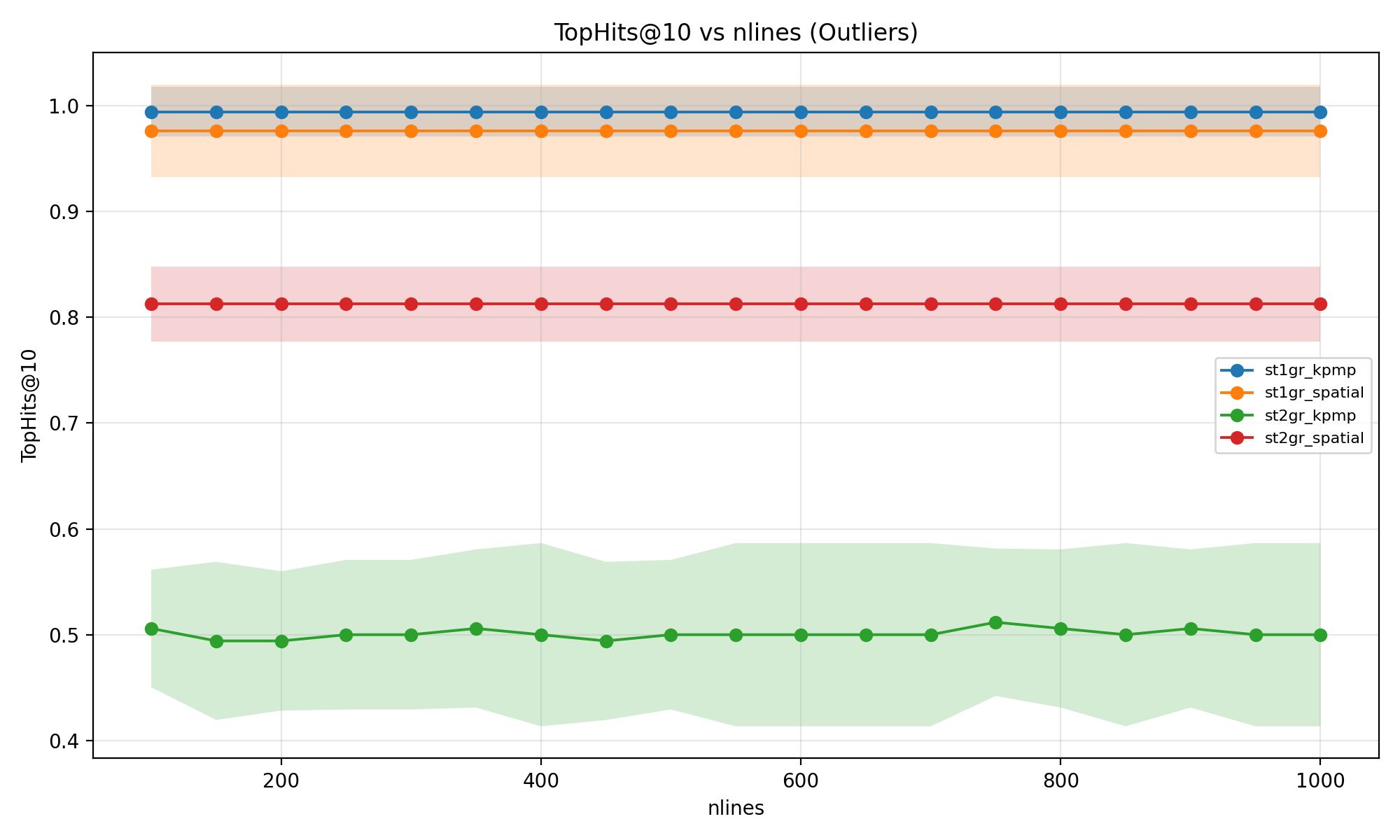} 
	\caption{Quality scores vs number of lines $|\linefam|$ for numerical experiments with outlier function values. As scores are roughly constant, this motivates keeping a moderate number of lines.}
	\label{fig:score-line-outliers}
\end{figure}

\begin{figure}[h!]
	\centering
	\includegraphics[width=12cm]{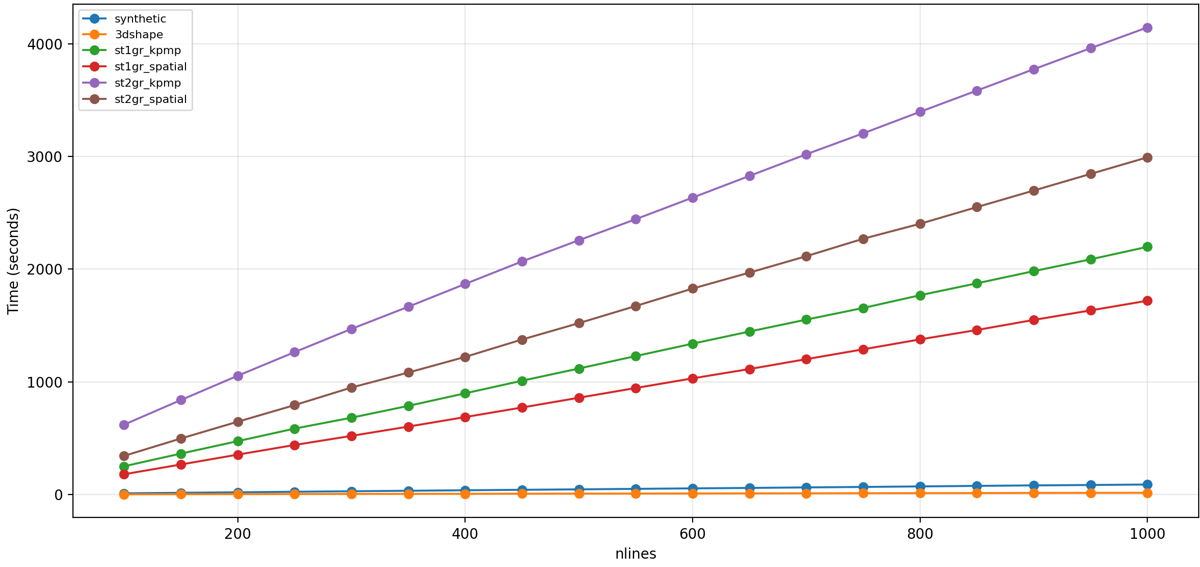} 
	\includegraphics[width=12cm]{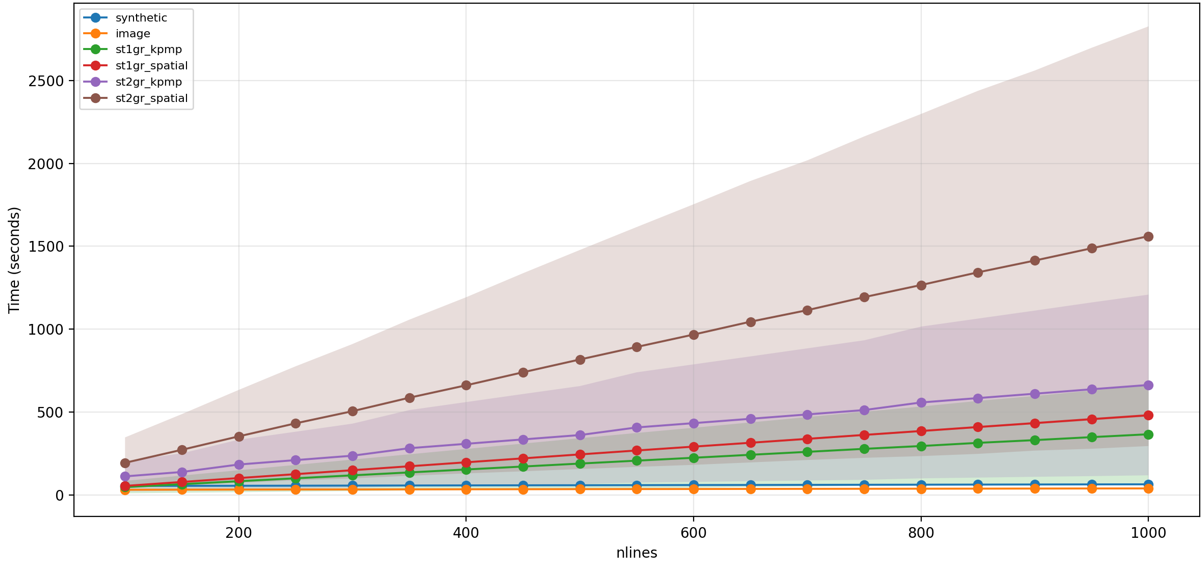} 
	\caption{\tomatomp\ running times for numerical  experiments with no graph parameter tuning \textbf{(top)} and outlier function values \textbf{(bottom)}.}
	\label{fig:time-line}
\end{figure}

\clearpage
\subsection{Additional plots}\label{app:add-plot}

\begin{figure}[h!]
	\centering
	\includegraphics[width=4cm]{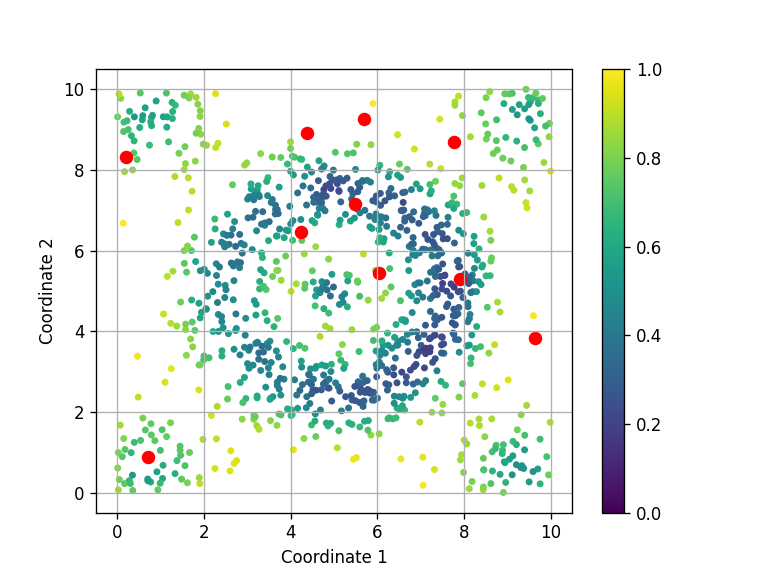}
	\includegraphics[width=4cm]{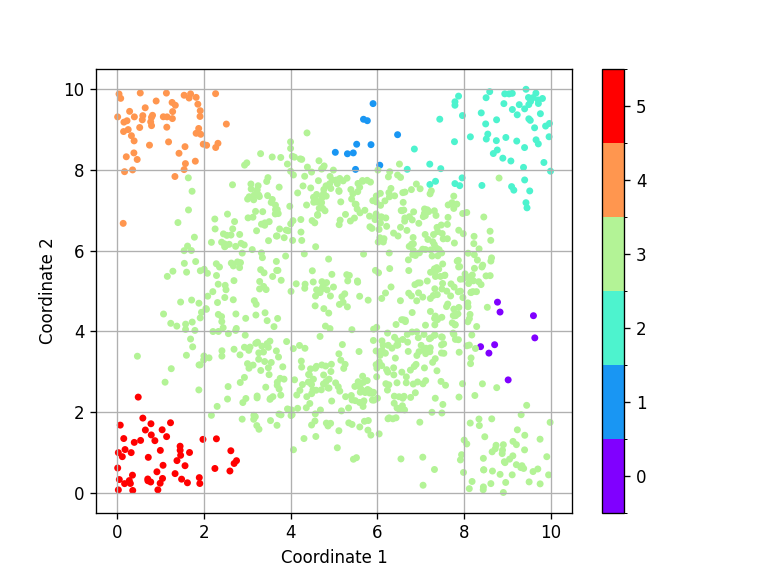}
	\includegraphics[width=4cm]{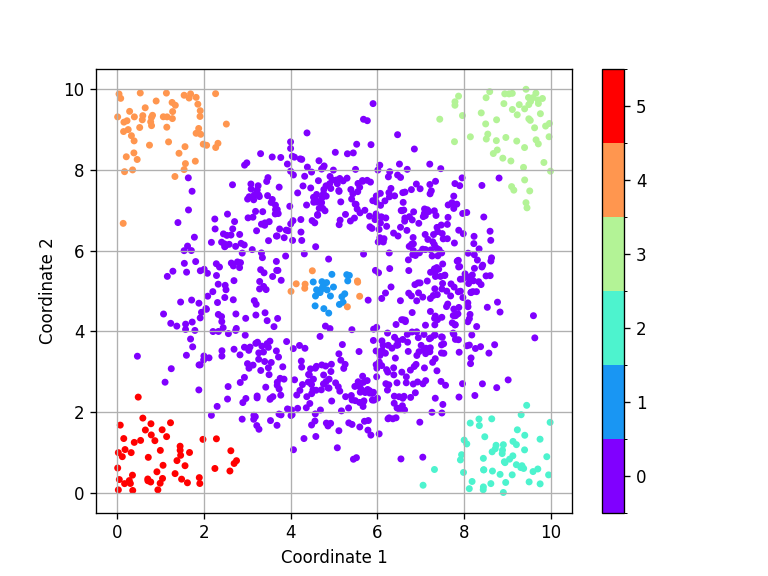}
	\includegraphics[width=4cm]{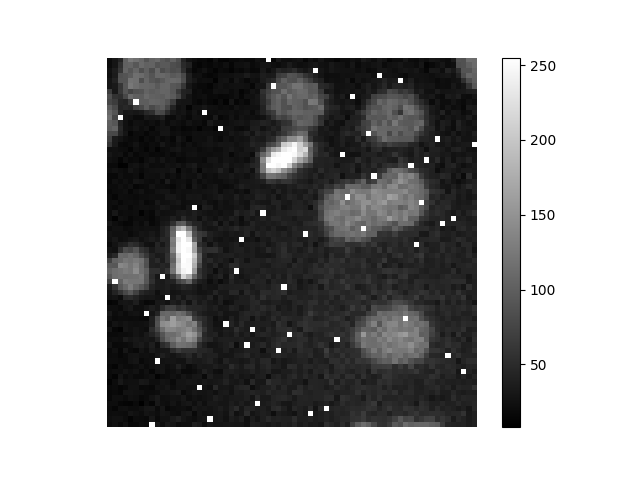}
	\includegraphics[width=4cm]{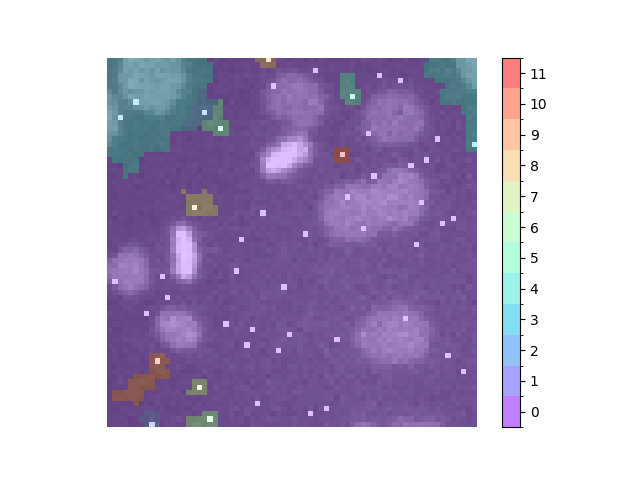}
	\includegraphics[width=4cm]{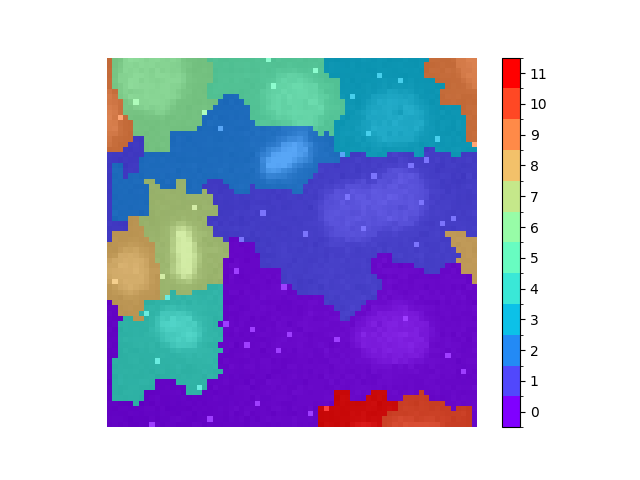}
	\caption{\textbf{(Left)} Synthetic dataset and morphological image plagued with outlier values. \textbf{(Middle)} \tomato\ clusterings are messed up by such outliers. \textbf{(Right)} \tomatomp\ clusters remain robust.}
	\label{fig:outlier-results}
\end{figure}

\begin{figure}[h!]
	\centering
	\includegraphics[width=9cm]{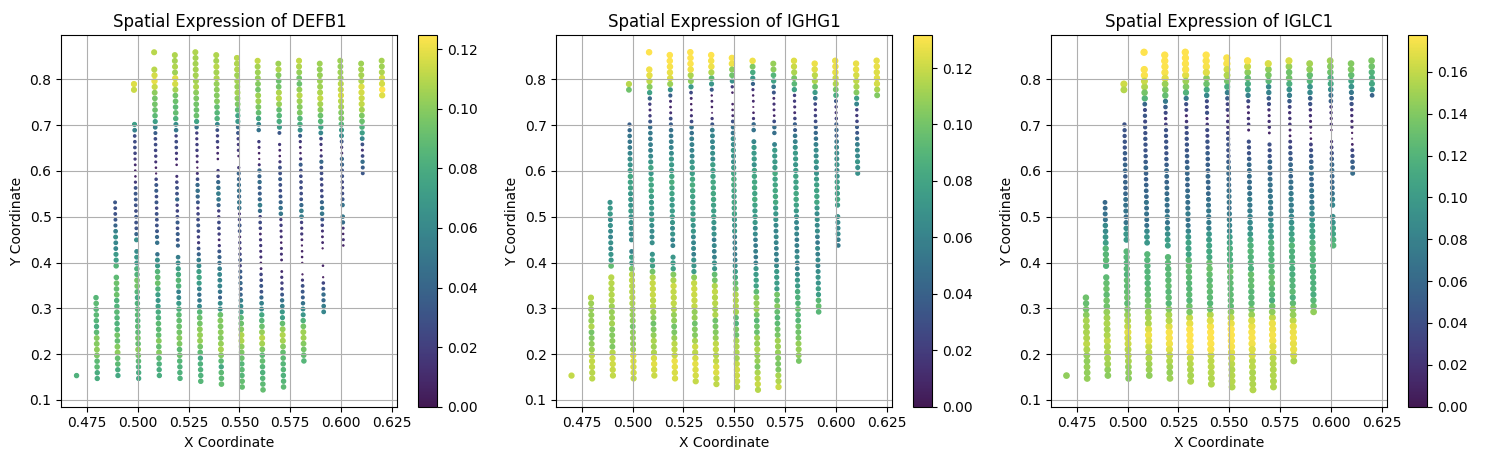}\ \ \ \ \ \ \ 
	\includegraphics[width=9cm]{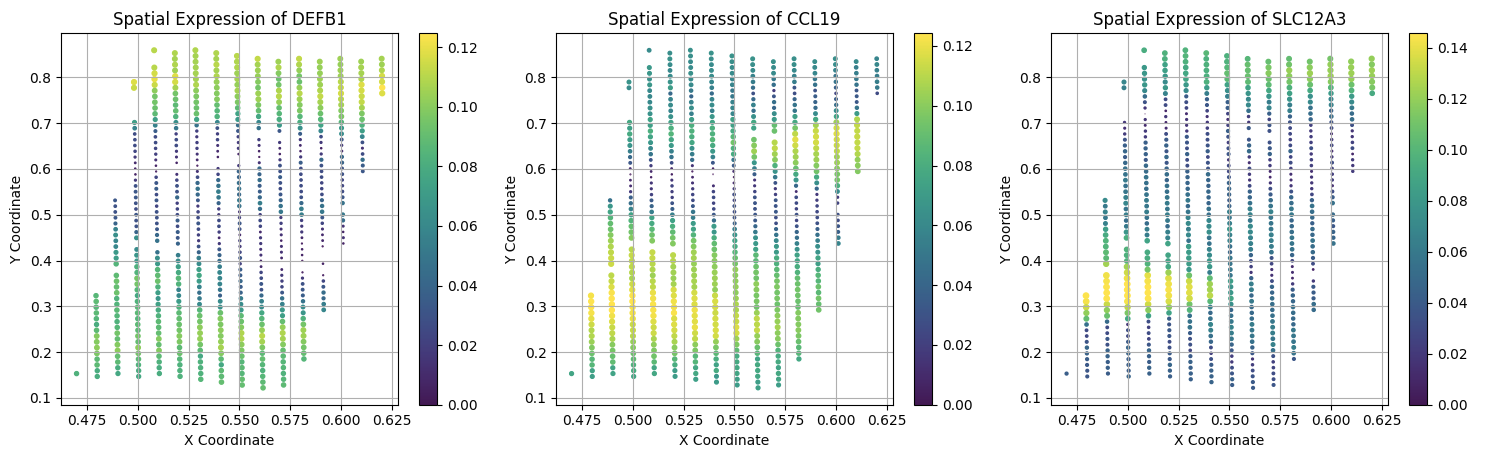}
	\caption{\textbf{(Top)} Top ranked triplet of genes for kpmp dataset.
	\textbf{(Bottom)} Lower ranked triplet of genes.}
	\label{fig:triplets-kpmp}
\end{figure}

\clearpage
\subsection{Additional tables}\label{app:add-table}

% Auto-generated by aggregate_results_to_latex.py
\begin{table}[h]
\centering
\scriptsize
\setlength{\tabcolsep}{3pt}
\caption{\label{tab:ari-metric-clus} ARI scores for numerical experiments with no graph parameter tuning.}
\resizebox{\textwidth}{!}{%
\begin{tabular}{|l|cccccc|cc|c|}
\hline
Dataset & KM(w/) & KM(w) & SC(w/) & SC(w) & HC(w/) & HC(w) & \tomato\ (min) & \tomato\ (mean) & \tomatomp \\ % (100) & AuToMATo (500) & AuToMATo (1000) \\
\hline
synthetic & 0.0764 & 0.4164 & 0.1431 & 0.0338 & -0.0019 & 0.0232 & 0.1780 & 0.7920 & \textbf{0.9551} \\ %& \textbf{0.9583} & 0.9551 \\
3dshape & 0.4901 & 0.5008 & 0.4776 & 0.4823 & 0.5003 & 0.4776 & 0.0017 & 0.3704 & \textbf{0.7333} \\ % & \textbf{0.7333} & \textbf{0.7333} \\
\hline
\end{tabular}%
}
\end{table}

% Auto-generated by aggregate_results_to_latex.py
\begin{table}[h]
\centering
\scriptsize
\setlength{\tabcolsep}{3pt}
\caption{\label{tab:pearson-metric-rank} Pearson correlation for numerical experiments with no graph parameter tuning.}
%\resizebox{\textwidth}{!}{%
\begin{tabular}{|l|c|cc|c|}
\hline
Dataset & HC(w) & \tomato\ (min) & \tomato\ (mean) & \tomatomp \\ %& AuToMATo (500, mode 1) & AuToMATo (1000, mode 1) & AuToMATo (100, mode 3) & AuToMATo (500, mode 3) & AuToMATo (1000, mode 3) & AuToMATo (100, mode 4) & AuToMATo (500, mode 4) & AuToMATo (1000, mode 4) & AuToMATo (100, mode 6) & AuToMATo (500, mode 6) & AuToMATo (1000, mode 6) \\
\hline
1-g kpmp & -0.5116 & 0.9100 & \textbf{0.9675} & 0.9621 \\ %& 0.9621 & 0.9621 & 0.9621 & 0.9621 & 0.9621 & 0.9621 & 0.9621 & 0.9621 & 0.9621 & 0.9621 & 0.9621 \\
1-g spat & -0.2582 & \textbf{1.0000} & \textbf{1.0000} & 0.9969 \\ %& 0.9969 & 0.9969 & 0.9969 & 0.9969 & 0.9969 & 0.9969 & 0.9969 & 0.9969 & 0.9969 & 0.9969 & 0.9969 \\
%st1gr kpmp (200) & -0.4754 & 0.9219 & \textbf{0.9738} & 0.9723 & 0.9723 & 0.9723 & 0.9723 & 0.9723 & 0.9723 & 0.9723 & 0.9723 & 0.9723 & 0.9723 & 0.9723 & 0.9723 \\
%st1gr spatial (200) & -0.2582 & \textbf{1.0000} & \textbf{1.0000} & 0.9969 & 0.9969 & 0.9969 & 0.9969 & 0.9969 & 0.9969 & 0.9969 & 0.9969 & 0.9969 & 0.9969 & 0.9969 & 0.9969 \\
2-g kpmp & -0.2139 & 0.9114 & \textbf{0.9771} & 0.0183 \\ %& -0.1191 & -0.1212 & 0.0183 & -0.1191 & -0.1212 & 0.0183 & -0.1191 & -0.1212 & 0.0183 & -0.1191 & -0.1212 \\
2-g spat & -0.5030 & \textbf{1.0000} & \textbf{1.0000} & 0.5643 \\ %& 0.5648 & 0.5654 & 0.5643 & 0.5648 & 0.5654 & 0.5643 & 0.5648 & 0.5654 & 0.5643 & 0.5648 & 0.5654 \\
%st2gr kpmp (20) & -0.2531 & 0.9303 & \textbf{0.9768} & 0.1500 & 0.0043 & -0.0082 & 0.1500 & 0.0043 & -0.0082 & 0.1500 & 0.0043 & -0.0082 & 0.1500 & 0.0043 & -0.0082 \\
%st2gr spatial (20) & -- & -- & -- & -- & -- & -- & -- & -- & -- & -- & -- & -- & -- & -- & -- \\
\hline
\end{tabular}%
%}
\end{table}

% Auto-generated by aggregate_results_to_latex.py
\begin{table}[h]
\centering
\scriptsize
\setlength{\tabcolsep}{3pt}
\caption{\label{tab:ari-outliers-clus} ARI scores for numerical experiments with outliers}
\resizebox{\textwidth}{!}{%
\begin{tabular}{|l|cccccc|c|c|}
\hline
Dataset & KM(w/) & KM(w) & SC(w/) & SC(w) & HC(w/) & HC(w) & \tomato & \tomatomp \\ % (100) & AuToMATo (500) & AuToMATo (1000) \\
\hline
synthetic & 0.1116 $\pm$ 0.0204 & 0.4132 $\pm$ 0.0027 & 0.1246 $\pm$ 0.0289 & 0.1042 $\pm$ 0.0104 & 0.0000 $\pm$ 0.0000 & 0.1068 $\pm$ 0.0312 & 0.5164 $\pm$ 0.1596 & \textbf{0.9542 $\pm$ 0.0124} \\% & \textbf{0.9543 $\pm$ 0.0123} & 0.9541 $\pm$ 0.0124 \\
image & 0.2083 $\pm$ 0.0117 & 0.1955 $\pm$ 0.0168 & -- & -- & 0.2054 $\pm$ 0.0122 & 0.1991 $\pm$ 0.0178 & 0.0566 $\pm$ 0.0438 & \textbf{0.9058 $\pm$ 0.0664} \\% & 0.9116 $\pm$ 0.0593 & \textbf{0.9119 $\pm$ 0.0620} \\
\hline
\end{tabular}%
}
\end{table}

% Auto-generated by aggregate_results_to_latex.py
\begin{table}[h!]
\centering
\scriptsize
\setlength{\tabcolsep}{3pt}
\caption{\label{tab:pearson-outliers-rank} Pearson correlation  for numerical experiments with outliers}
%\resizebox{\textwidth}{!}{%
\begin{tabular}{|l|cc|c|c|}
\hline
Dataset & HC(w/) & HC(w) & \tomato & \tomatomp \\ %AuToMATo (100, mode 1) & AuToMATo (500, mode 1) & AuToMATo (1000, mode 1) & AuToMATo (100, mode 3) & AuToMATo (500, mode 3) & AuToMATo (1000, mode 3) & AuToMATo (100, mode 4) & AuToMATo (500, mode 4) & AuToMATo (1000, mode 4) & AuToMATo (100, mode 6) & AuToMATo (500, mode 6) & AuToMATo (1000, mode 6) \\
\hline
1-g kpmp & -0.4294 $\pm$ 0.0661 & -0.9376 $\pm$ 0.0030 & 0.4131 $\pm$ 0.0516 & \textbf{0.9938 $\pm$ 0.0215} \\ % & \textbf{0.9939 $\pm$ 0.0215} & \textbf{0.9939 $\pm$ 0.0215} & 0.3283 $\pm$ 0.0911 & 0.3913 $\pm$ 0.1944 & 0.4355 $\pm$ 0.1271 & -0.1773 $\pm$ 0.0937 & -0.3751 $\pm$ 0.1295 & -0.1584 $\pm$ 0.1335 & -0.2386 $\pm$ 0.1299 & -0.1529 $\pm$ 0.1222 & -0.0637 $\pm$ 0.1201 \\
1-gr spat & -0.7931 $\pm$ 0.0044 & -0.9669 $\pm$ 0.0051 & 0.9365 $\pm$ 0.0412 & \textbf{0.9797 $\pm$ 0.0100} \\ % & 0.9796 $\pm$ 0.0100 & 0.9796 $\pm$ 0.0100 & 0.9442 $\pm$ 0.0355 & 0.9442 $\pm$ 0.0355 & 0.9442 $\pm$ 0.0355 & 0.9794 $\pm$ 0.0101 & 0.9798 $\pm$ 0.0099 & \textbf{0.9798 $\pm$ 0.0098} & 0.9439 $\pm$ 0.0357 & 0.9441 $\pm$ 0.0355 & 0.9442 $\pm$ 0.0355 \\
%st1gr kpmp (200) & -0.4718 $\pm$ 0.0554 & -0.8757 $\pm$ 0.0035 & 0.4854 $\pm$ 0.0605 & 0.9977 $\pm$ 0.0022 & \textbf{0.9977 $\pm$ 0.0022} & \textbf{0.9977 $\pm$ 0.0022} & 0.4007 $\pm$ 0.0314 & 0.4871 $\pm$ 0.0445 & 0.5230 $\pm$ 0.0466 & -0.1235 $\pm$ 0.0431 & -0.4930 $\pm$ 0.0552 & -0.3892 $\pm$ 0.0517 & -0.2601 $\pm$ 0.0325 & -0.2901 $\pm$ 0.0673 & -0.1932 $\pm$ 0.0733 \\
%st1gr spatial (200) & -0.7933 $\pm$ 0.0045 & -0.9670 $\pm$ 0.0053 & 0.9386 $\pm$ 0.0416 & 0.9798 $\pm$ 0.0103 & 0.9798 $\pm$ 0.0103 & 0.9798 $\pm$ 0.0103 & 0.9462 $\pm$ 0.0357 & 0.9462 $\pm$ 0.0357 & 0.9462 $\pm$ 0.0357 & 0.9796 $\pm$ 0.0104 & 0.9799 $\pm$ 0.0102 & \textbf{0.9799 $\pm$ 0.0102} & 0.9459 $\pm$ 0.0359 & 0.9461 $\pm$ 0.0357 & 0.9462 $\pm$ 0.0357 \\
2-g kpmp & -0.1657 $\pm$ 0.0933 & \textbf{-0.6113 $\pm$ 0.0030} & 0.1287 $\pm$ 0.1406 & 0.1442 $\pm$ 0.0615 \\ %& 0.1603 $\pm$ 0.0637 & \textbf{0.1612 $\pm$ 0.0651} & 0.1545 $\pm$ 0.1027 & 0.0625 $\pm$ 0.0866 & 0.0433 $\pm$ 0.1338 & -0.0111 $\pm$ 0.1237 & -0.1467 $\pm$ 0.1165 & -0.2357 $\pm$ 0.1405 & -0.0157 $\pm$ 0.1210 & -0.1500 $\pm$ 0.1119 & -0.1453 $\pm$ 0.0846 \\
2-g spat & -0.3536 $\pm$ 0.0239 & \textbf{-0.9339 $\pm$ 0.0024} & 0.7672 $\pm$ 0.0926 & 0.5994 $\pm$ 0.0308 \\ % & 0.5954 $\pm$ 0.0337 & 0.5779 $\pm$ 0.0266 & 0.6600 $\pm$ 0.1186 & 0.1349 $\pm$ 0.2733 & 0.2104 $\pm$ 0.3142 & 0.5922 $\pm$ 0.0641 & 0.5753 $\pm$ 0.0669 & 0.5796 $\pm$ 0.0462 & 0.5967 $\pm$ 0.1227 & 0.1965 $\pm$ 0.2570 & 0.2713 $\pm$ 0.2703 \\
%st2gr kpmp (20) & -0.2762 $\pm$ 0.0594 & -0.5409 $\pm$ 0.0002 & 0.3020 $\pm$ 0.0920 & 0.3396 $\pm$ 0.0173 & \textbf{0.3481 $\pm$ 0.0168} & 0.3329 $\pm$ 0.0061 & 0.2166 $\pm$ 0.0714 & 0.1562 $\pm$ 0.0494 & 0.1809 $\pm$ 0.0567 & 0.0063 $\pm$ 0.0360 & -0.2379 $\pm$ 0.1119 & -0.2278 $\pm$ 0.1004 & -0.0431 $\pm$ 0.0303 & -0.2424 $\pm$ 0.1115 & -0.1747 $\pm$ 0.0710 \\
%st2gr spatial (20) & -0.4732 $\pm$ 0.0004 & -0.9643 $\pm$ 0.0002 & \textbf{0.8054 $\pm$ 0.0284} & 0.7150 $\pm$ 0.0060 & 0.7115 $\pm$ 0.0047 & 0.7094 $\pm$ 0.0040 & 0.7641 $\pm$ 0.0161 & 0.3211 $\pm$ 0.0879 & 0.4416 $\pm$ 0.1440 & 0.6652 $\pm$ 0.0345 & 0.6619 $\pm$ 0.0297 & 0.6820 $\pm$ 0.0108 & 0.7048 $\pm$ 0.1247 & 0.2924 $\pm$ 0.0690 & 0.4465 $\pm$ 0.1522 \\
\hline
\end{tabular}%
%}
\end{table}

%\clearpage
%\input{checklist}

\end{document}